\documentclass[acmsmall]{acmart}

\usepackage{multirow,multicol}
\usepackage{subfigure}
\usepackage{color}
\AtBeginDocument{%
  \providecommand\BibTeX{{%
    \normalfont B\kern-0.5em{\scshape i\kern-0.25em b}\kern-0.8em\TeX}}}

\begin{document}

\title{Trustworthy AI: A Computational Perspective}


\author{Haochen Liu}
\authornote{Both authors contributed equally to this work.}
\affiliation{%
  \institution{Michigan State University}
  \city{East Lansing}
  \state{MI}
  \country{USA}}
\email{liuhaoc1@msu.edu}

\author{Yiqi Wang}
\authornotemark[1]
\affiliation{%
  \institution{Michigan State University}
  \city{East Lansing}
  \state{MI}
  \country{USA}}
\email{wangy206@msu.edu}

\author{Wenqi Fan}
\affiliation{%
  \institution{The Hong Kong Polytechnic University}
  \country{Hong Kong}}
\email{wenqifan03@gmail.com}

\author{Xiaorui Liu}
\affiliation{%
  \institution{Michigan State University}
  \city{East Lansing}
  \state{MI}
  \country{USA}}
\email{xiaorui@msu.edu}

\author{Yaxin Li}
\affiliation{%
  \institution{Michigan State University}
  \city{East Lansing}
  \state{MI}
  \country{USA}}
\email{liyaxin1@msu.edu}

\author{Shaili Jain}
\affiliation{%
  \institution{Twitter}
  \country{USA}}
\email{shailij@gmail.com}

\author{Yunhao Liu}
\affiliation{%
  \institution{Tsinghua University}
  \country{China}}
\email{yunhao@tsinghua.edu.cn}

\author{Anil K. Jain}
\affiliation{%
  \institution{Michigan State University}
  \city{East Lansing}
  \state{MI}
  \country{USA}}
\email{jain@cse.msu.edu}

\author{Jiliang Tang}
\affiliation{%
  \institution{Michigan State University}
  \city{East Lansing}
  \state{MI}
  \country{USA}}
\email{tangjili@msu.edu}

\renewcommand{\shortauthors}{Liu and Wang, et al.}

\begin{abstract}
In the past few decades, artificial intelligence (AI) technology has experienced swift developments, changing everyone’s daily life and profoundly altering the course of human society. The intention behind developing AI was and is to benefit humans by reducing labor, increasing everyday conveniences, and promoting social good. However, recent research and AI applications indicate that AI can cause unintentional harm to humans by, for example, making unreliable decisions in safety-critical scenarios or undermining fairness by inadvertently discriminating against a group or groups. Consequently, trustworthy AI has recently garnered increased attention regarding the need to avoid the adverse effects that AI could bring to people, so people can fully trust and live in harmony with AI technologies.

A tremendous amount of research on trustworthy AI has been conducted and witnessed in recent years. In this survey, we present a comprehensive appraisal of trustworthy AI from a computational perspective to help readers understand the latest technologies for achieving trustworthy AI. Trustworthy AI is a large and complex subject, involving various dimensions. In this work, we focus on six of the most crucial dimensions in achieving trustworthy AI: (i) Safety \& Robustness, (ii) Nondiscrimination \& Fairness, (iii) Explainability, (iv) Privacy, (v) Accountability \& Auditability, and (vi) Environmental Well-being. For each dimension, we review the recent related technologies according to a taxonomy and summarize their applications in real-world systems. We also discuss the accordant and conflicting interactions among different dimensions and discuss potential aspects for trustworthy AI to investigate in the future.
\end{abstract}



\maketitle

\section{Introduction}

Artificial intelligence (AI), a science that studies and develops the theory, methodology, technology, and application systems for simulating, extending, and expanding human intelligence, has brought revolutionary impact to modern human society. From a micro view, AI plays an irreplaceable role in many aspects of our lives. Modern life is filled with interactions with AI applications: from unlocking a cell phone with face ID, talking to a voice assistant, and buying products recommended by e-commerce platforms; from a macro view, AI creates great economic outcomes. The Future of Jobs Report 2020 from the World Economic Forum \cite{world2020future} predicts that AI will create 58 million new jobs in five years. By 2030, AI is expected to produce extra economic profits of 13 trillion U.S. dollars, which contribute 1.2\% annual growth to the GDP of the whole world \cite{bughin2018notes}. However, along with their rapid and impressive development, AI systems have also exposed their untrustworthy sides. For example, safety-critical AI systems are shown to be vulnerable to adversarial attacks. Deep image recognition systems in autonomous vehicles could fail to distinguish road signs modified by malicious attackers \cite{xu2020adversarial}, posing great threat to passenger safety. In addition, AI algorithms can cause bias and unfairness. Online AI chatbots could produce indecent, racist, and sexist content \cite{wolf2017we} that offends users and has a negative social impact. Moreover, AI systems carry the risk of disclosing user privacy and business secrets. Hackers can take advantage of the feature vectors produced by an AI model to reconstruct private input data, such as fingerprints \cite{al2019privacy}, thereby leaking a user’s sensitive information. These vulnerabilities can make existing AI systems unusable and can cause severe economic and security consequences. Concerns around trustwor- thiness have become a huge obstacle for AI to overcome to advance as a field, become more widely adopted, and create more economic value. Hence, how to build trustworthy AI systems has become a focal topic in both academia and industry.

In recent years, a large body of literature on trustworthy AI has emerged. With the increasing demand for building trustworthy AI, it is imperative to summarize existing achievements and discuss possible directions for future research. In this survey, we provide a comprehensive overview of trustworthy AI to help newcomers attain a basic understanding of what makes an AI system trustworthy and to help veterans track the latest progress in the field. We clarify the definition of trustworthy AI and introduce six key dimensions of it. For each dimension, we present its concepts and taxonomies and review representative algorithms. We also introduce possible interactions among different dimensions and discuss other potential issues around trustworthy AI that have not yet drawn sufficient attention. In addition to definitions and concepts, our survey focuses on the specific computational solutions for realizing each dimension of trustworthy AI. This perspective makes it distinct from some extant related works, such as a government guideline \cite{smuha2019eu}, which suggests how to build a trustworthy AI system in the form of laws and regulations, or reviews \cite{brundage2020toward,thiebes2020trustworthy}, which discuss the realization of trustworthy AI from a high-level, nontechnical perspective.

\begin{figure}[ht]
\includegraphics[scale=0.5]{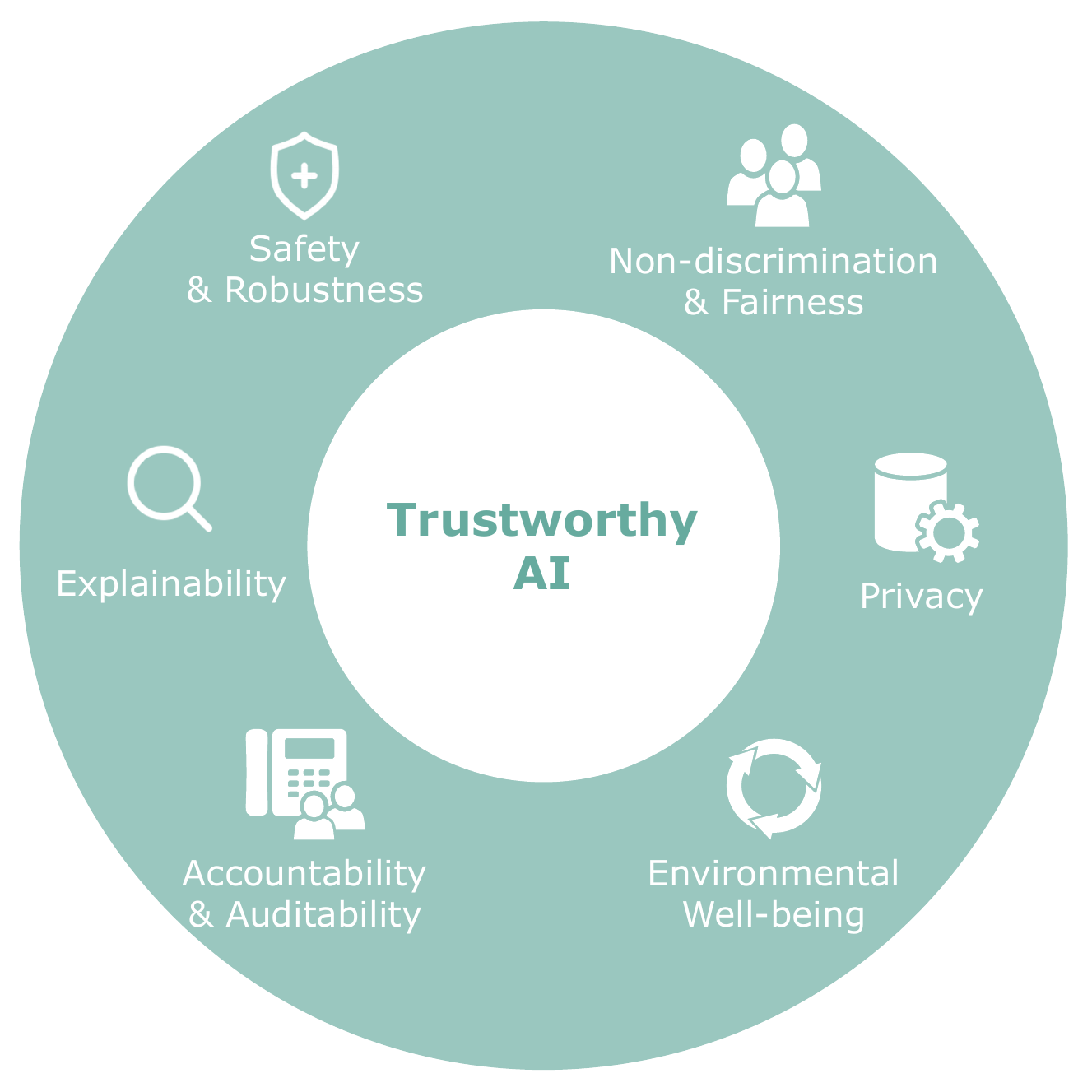}
\caption{Six key dimensions of trustworthy AI.}
\label{fig:intro}
\end{figure}

According to a recent ethics guideline for AI provided by the European Union (EU) \cite{smuha2019eu}, a trustworthy AI system should meet four ethical principles: respect for human autonomy, prevention of harm, fairness, and explicability. Based on these four principles, AI researchers, practitioners, and governments propose various specific dimensions for trustworthy AI \cite{smuha2019eu,thiebes2020trustworthy,brundage2020toward}. In this survey, we focus on six important and concerning dimensions that have been extensively studied. As shown in Figure \ref{fig:intro}, they are \textbf{Safety \& Robustness, Non-discrimination \& Fairness, Explainability, Privacy, Auditability \& Accountability, and Environmental Well-Being}.

The remaining survey is organized as follows. In section \ref{sec:concept}, we articulate the definition of trustworthy AI and provide various definitions of it to help readers understand how a trustworthy AI system is defined by researchers from such different disciplines as computer science, sociology, law, and business. We then distinguish the concept of trustworthy AI from several related concepts, such as ethical AI and responsible AI.

In section \ref{sec:safety}, we detail the dimension of \textbf{Safety \& Robustness}, which requires an AI system to be robust to the noisy perturbations of inputs and to be able to make secure decisions. In recent years, numerous studies have shown that AI systems, especially those that adopt deep learning models, can be very sensitive to intentional or unintentional inputs perturbations, posing huge risks to safety-critical applications. For example, as described before, autonomous vehicles can be fooled by altered road signs. Additionally, spam detection models can be fooled by emails with well-designed text \cite{barreno2010security}. Thus, spam senders can take advantage of this weakness to make their emails immune to the detection system, which would cause a bad user experience.

It has been demonstrated that AI algorithms can learn human discriminations through provided training examples and make unfair decisions. For example, some face recognition algorithms have difficulty detecting faces of African Americans \cite{rose2010face} or misclassifying them as gorillas \cite{howard2018ugly}. Moreover, voice dictation software typically performs better at recognizing a voice from a male than that from a female \cite{rodger2004field}. In section \ref{sec:fairness}, we introduce the dimension of \textbf{Non-discrimination \& Fairness} in which an AI system is expected to avoid unfair bias toward certain groups or individuals.

In section \ref{sec:interpret}, we discuss the dimension of \textbf{Explainability}, which suggests that the AI’s decision mechanism system should be able to be explained to stakeholders (who should be able to understand the explanation). For example, AI techniques have been used for disease diagnosis based on the symptoms and physical features of a patient \cite{sajda2006machine}. In such cases, a black-box decision is not acceptable. The inference process should be transparent to doctors and patients to ensure that the diagnosis is exact in every detail.

Researchers have found that some AI algorithms can store and expose users’ personal information. For example, dialogue models trained on the human conversation corpus can remember sensitive information, like credit card numbers, which can be elicited by interacting with the model \cite{henderson2018ethical}. In section \ref{sec:privacy}, we present the dimension of \textbf{Privacy}, which requires an AI system to avoid leaking any private information.

In section \ref{sec:account}, we describe the dimension of \textbf{Auditability \& Accountability}, which expects that an AI system be assessed by a third party and, when necessary, assign responsibility for an AI failure, especially in critical applications \cite{smuha2019eu}.

Recently, the environmental impacts of AI systems have drawn people’s attention, since some large AI systems consume great amounts of energy. As a mainstream AI technology, deep learning is moving toward pursuing larger models and more parameters. Accordingly, more storage and computational resources are consumed. A study \cite{strubell2019energy} shows that training a BERT model \cite{devlin2019bert} costs a carbon emission of around 1,400 pounds of carbon dioxide, which is comparable to that of a round trip trans-America flight. Therefore, an AI system should be sustainable and environmentally friendly. In section \ref{sec:environment}, we review the dimension of \textbf{Environmental Well-Being}.

In section \ref{sec:relation}, we discuss the interactions among the different dimensions. Recent studies have demonstrated that there are accordance and conflicts among different dimensions of trustworthy AI \cite{smuha2019eu,whittlestone2019role}. For example, the robustness and explainability of deep neural networks are tightly connected and robust models tend to be more interpretable \cite{tsipras2018robustness,etmann2019connection} and vice versa \cite{noack2021empirical}. Moreover, it is shown that in some cases, a trade-off exists between robust ness and privacy. For instance, adversarial defense approaches can make a model more vulnerable to membership inference attacks, which increases the risk of training data leakage \cite{song2019privacy}.

In addition to the aforementioned six dimensions, there are more dimensions of trustworthy AI, such as human agency and oversight, creditability, etc. Although these additional dimensions are as important as the six dimensions considered in this article, they are in earlier stages of development with limited literature, especially for computational methods. Thus, in section \ref{sec:future}, we discuss these dimensions of trustworthy AI as future directions needing dedicated research efforts.
\section{Concepts and Definitions}
\label{sec:concept}

The word ``trustworthy'' is noted to mean ``worthy of trust of confidence; reliable, dependable'' in the Oxford English Dictionary or ``able to be trusted'' in the Dictionary of Cambridge. Trustworthy descends from the word trust, which is described as the ``firm belief in the reliability, truth, or ability of someone or something'' in the Oxford English Dictionary or the ``belief that you can depend on someone or something'' in the Dictionary of Cambridge. Broadly speaking, trust is a widespread notion in human society, which lays the important foundation for the sustainable development of human civilization. Strictly speaking, some potential risks always exist in our external environment because we cannot completely control people and other entities \cite{mayer1995integrative,thiebes2020trustworthy}. It is our trust in these parties that allows us to put ourselves at potential risk to continue interacting with them willingly \cite{lee2004trust}. Trust is necessary among people. It forms the basis of a good relationship and is necessary for people to live happily and to work efficiently together. In addition, trust is also vital between humans and technology. Without trust, humans would not be willing to utilize technology, which would undoubtedly impede its advancement and prevent people from enjoying the conveniences it brings. Therefore, for a win-win situation between humans and technology, it is necessary to guarantee that the technology is trustworthy so people can build trust in it.

The term ``artificial intelligence'' got its name from a workshop at a 1956 Dartmouth conference \cite{buchanan2005very,mccarthy2006proposal}. Although there are numerous definitions for AI \cite{kok2009artificial}, AI generally denotes a program or system that is able to cope with a real-world problem with human-like reasoning, for example, in the field of image recognition within AI, which uses deep learning networks to recognize objects or people within images \cite{rawat2017deep}. The past few decades have witnessed rapid and impressive development of AI; there are tons of breakthroughs happening in every corner of this field \cite{russell2002artificial,sze2017efficient}. Furthermore, with the rapid development of big data and computational resources, AI has been broadly applied to many aspects of human life, including economics, healthcare, education, transportation, and so on, where it has revolutionized industries and achieved numerous feats. Considering the important role AI plays in modern society, it is necessary to make AI trustworthy so that humans can rely on it with minimal concern regarding its potential harm. Trust is essential in allowing the potential of AI to be fully realized – and humans to fully enjoy its benefits and convenience \cite{ec2019independent}.

\begin{table}[ht]
	\centering
	\caption{A summary of principles for Trustworthy AI from different perspectives.}
	\label{table:concept}
	\begin{tabular}{@{}|c|c|@{}}
		\hline
		\textbf{Perspective} & \textbf{Principles}\\
		\hline
		\multirow{2}{*}{Technical} & Accuracy, Robustness, \\ 
		&  Explainability \\  \hline
		\multirow{2}{*}{User} & Availability, Usability, \\
		&  Safety, Privacy, Autonomy  \\  \hline
		\multirow{2}{*}{Social} & Law-abiding, Ethical, Fair, \\  
		&  Accountable, Environmental-friendly   \\  \hline
	\end{tabular}
	\vspace{-3mm}
\end{table}

Due to its importance and necessity, trustworthy AI has drawn increasing attention, and there are numerous discussions and debates over its definition and extension \cite{ec2019independent}. In this survey, we define trustworthy AI as \textit{programs and systems built to solve problems like a human, which bring benefits and convenience to people with no threat or risk of harm}. We further define trustworthy AI from the following three perspectives: the technical perspective, the user perspective, and the social perspective. An overall description of these perspectives is summarized in Table \ref{table:concept}.
\begin{itemize}
\item \textbf{From a technical perspective}, trustworthy AI is expected to show the properties of accuracy, robustness, and explainability. Specifically, AI programs or systems should generate accurate output consistent with the ground truth as much as possible. This is also the first and most basic motivation for building them. Additionally, AI programs or systems should be robust to changes that should not affect their outcome. This is very important, since real environments where AI systems are deployed are usually very complex and volatile. Last, but not least, trustworthy AI must allow for explanation and analysis by humans, so that potential risks and harm can be minimized. In addition, trustworthy AI should be transparent so people can better understand its mechanism.

\item \textbf{From a user's perspective}, trustworthy AI should possess the properties of availability, usability, safety, privacy, and autonomy. Specifically, AI programs or systems should be available for people whenever they need them, and these programs or systems should be easy to use for people with different backgrounds. More importantly, no AI programs or systems are expected to harm any people under any conditions, and to always put the safety of users as the first priority. In addition, trustworthy AI should protect the privacy of all users. It should deal with data storage very carefully and seriously. Last, but not least, the autonomy of trustworthy AI should always be under people’s control. In other words, it is always a human’s right to grant an AI system any decision-making power or to withdraw that power at any time.

\item \textbf{From a social perspective}, trustworthy AI should be law-abiding, ethical, fair, accountable, and environmentally friendly. Specifically, AI programs or systems should operate in full compliance with all relevant laws and regulations and comply with the ethical principles of human society. Importantly, trustworthy AI should show nondiscrimination toward people from various backgrounds. It should guarantee justice and fairness among all users. Also, trustworthy AI should be accountable, which means it is clear who is responsible for each part of the AI system. Lastly, for the sustainable development and long-term prosperity of our civilization, AI programs and systems should be environmentally friendly. For example, they should limit energy consumption and cause minimal pollution.
\end{itemize}

Note that the above properties of the three perspectives are not independent of each other. Instead, they complement and reinforce each other.

There have been numerous terminologies related to AI proposed recently, including ethical AI, beneficial AI, responsible AI, explainable AI, fair AI, and so on. These terminologies share some overlap and distinction with trustworthy AI in terms of the intention and extension of the concept.

Next, we briefly describe some related terminologies to help enhance the understanding of trustworthy AI.

\begin{itemize}
    \item \textbf{Ethical AI} \cite{floridi2018ai4peoplean}: An ethical framework of AI that specifies five core principles, including beneficence, nonmaleficence, autonomy, justice, and explicability. Additionally, 20 specific action points from four categories have been proposed to ensure continuous and effective efforts. They are assessment, development, incentivization, and support.
    
    \item \textbf{Beneficial AI} \cite{future2017asilomar}: AI has undoubtedly brought people count- less benefits, but to gain sustainable benefits from AI, 23 principles have been proposed in conjunction with the 2017 Asilomar conference. These principles are based on three aspects: research issues, ethics and values, and longer-term issues.
    
    \item \textbf{Responsible AI} \cite{MonresponsibleAI,CNGCresponsibleAI}: A framework for the development of responsible AI consists of 10 ethical principles: well-being, respect for autonomy, privacy and intimacy, solidarity, democratic participation, equity, diversity inclusion, prudence, responsibility, and sustainable development. The Chinese National Governance Committee for the New Generation Artificial Intelligence has proposed a set of governance principles to promote the healthy and sustainable development of responsible AI. Eight principles have been listed as follows: harmony and human-friendliness, fairness and justice, inclusion and sharing, respect for privacy, safety and controllability, shared responsibility, open and collaborative, and agile governance.
    
    \item \textbf{Explainable AI} \cite{adadi2018peeking}: The basic aim of explainable AI is to open up the "black box" of AI, to offer a trustworthy explanation of AI to users. It also aims to propose more explainable AI models, which can provide promising model performance and can be explained in nontechnical terms at the same time, so that users can fully trust them and take full advantage of them.
    
    \item \textbf{Fair AI} \cite{zou2018ai}: Because AI is designed by humans and data plays a key role in most AI models, it is easy for AI to inherit some bias from its creators or input data. Without proper guidance and regulations, AI could be biased and unfair toward a certain group or groups of people. Fair AI denotes AI that shows no discrimination toward people from any group. Its output should have little correlation with the traits of individuals, such as gender and ethnicity.

\end{itemize}

\begin{figure}[h]
\centering
\includegraphics[scale=0.45]{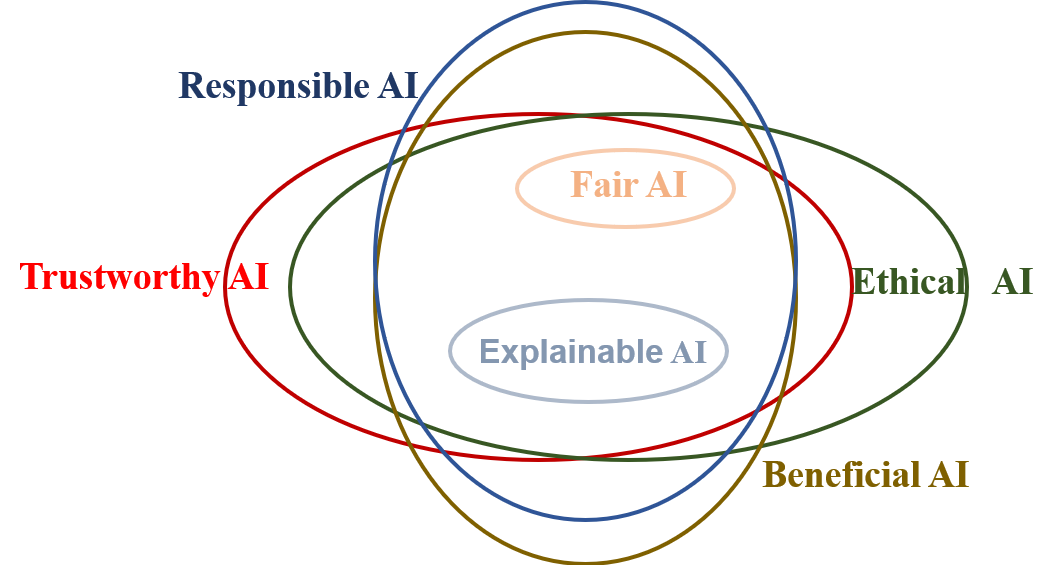}
\caption{The relation between trustworthy AI and related concepts.}
\label{fig:concept-ven}
\end{figure}

Overall, trustworthy AI has very rich connotation and can be interpreted in several perspectives. It contains the concepts of many existing terminologies, including fair AI, explainable AI, and so on. Huge overlaps also exist among the proposed concept of trustworthy AI and the concepts of ethical AI, beneficial AI, and responsible AI. All of them aim at building reliable AI that sustainably benefits human society. Nonetheless, some differences exist among these concepts since they are proposed from different perspectives. For example, in beneficial AI and responsible AI, there are also some principles and requirements for the users of AI and governors while in the proposed trustworthy AI, the main focus is on the principles of the AI technology itself. Figure \ref{fig:concept-ven} illustrates the relationships among these concepts.

\section{Safety \& Robustness}
\label{sec:safety}

A trustworthy AI system should achieve stable and sustained high accuracy under different circumstances. Specifically, a model should be robust to small perturbations since real-world data contains diverse types of noise. In recent years, many studies have shown that machine learning (ML) models can be fooled by small, designed perturbations, namely, adversarial perturbations \cite{szegedy2013intriguing, madry2017towards}. From traditional machine classifiers \cite{biggio2013evasion} to deep learning models, like CNN \cite{szegedy2013intriguing}, GNN \cite{scarselli2008graph}, or RNN \cite{zaremba2014recurrent}, none of the models are sufficiently robust to such perturbations. This raises huge concerns when ML models are applied to safety-critical tasks, such as authentication \cite{chen2017targeted}, auto-driving \cite{sitawarin2018darts}, recommendation \cite{fan2021attacking,fang2018poisoning}, AI healthcare \cite{finlayson2019adversarial}, etc. To build safe and reliable ML models, studying adversarial examples and the underlying reasons is urgent and essential.

In this section, we aim to introduce the concepts of robustness, how attackers design threat models, and how we develop different types of defense strategies. We first introduce the concepts for adversarial robustness. Then, we provide more details by introducing the taxonomy and giving examples for each category. We then discuss different adversarial attacks and defense strategies and introduce representative methods. Next, we introduce how adversarial robustness issues affect real-world AI systems. We also present existing related tools and surveys to help readers attain easy access. Finally, we demonstrate some potential future directions in adversarial robustness.

\subsection{Concepts and Taxonomy}
\label{sec:rob_con}

In this subsection, we briefly describe the commonly used and fundamental concepts in AI robustness to illustrate an overview of adversarial attacks and defenses and introduce the taxonomy based on these concepts.

\subsubsection{Threat Models}
An adversarial threat model, which can be denoted as an adversarial attack, is a process that an attacker uses to try to break the performance of ML models with fake training or test examples. The existence of adversarial attacks could lead to serious security concerns in a wide range of ML applications. Attackers use many different types of strategies to achieve their goal, hence threat models can be categorized into different types. In this subsection, we introduce different categories of threat models from different aspects of the ML program, including when the attack happens, what knowledge an attacker can access, and what the adversary’s goal is.

\begin{figure}[h]
\centering
\includegraphics[scale=0.45]{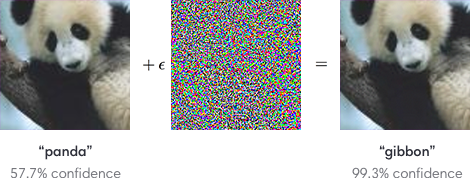}
\caption{An example of evasion attack. (Image Credit: \cite{goodfellow2014explaining})}
\label{fig:evasion}
\end{figure}

\textbf{Poisoning Attacks vs. Evasion Attacks.} Whether an attack is evasion or poisoning depends on whether attackers modify the training or test samples. A \textit{poisoning attack} occurs when attackers add fake samples into the training set of a classification model. These fake samples are designed intentionally to train a bad classifier, which achieves overall bad performance \citep{biggio2012poisoning} or gives wrong predictions on certain test samples \citep{zugner2018adversarial}. This type of attack can happen when the adversary has access to the training data and is a realistic safety concern. For example, training data for an online shopping recommendation system is often collected from web users where attackers may exist. A special case of a poisoning attack is a backdoor attack in which a trigger known only by the attacker is added to the training examples. This trigger is assigned with a target wrong label at the same time. The test samples with such a trigger are classified as the target label, which would cause severe harm for authentication systems, such as face recognition \cite{chen2017targeted}. An \textit{evasion attack} happens in the test phase. Given a well-trained classifier, attackers aim to design small perturbations for test samples in order to elicit wrong predictions from a victim model. From Figure \ref{fig:evasion}, we can see that the image of panda can be correctly classified by the model, while the perturbed version will be classified as a gibbon.

\textbf{White-box attacks vs. Black-box attacks.} According to the adversary’s knowledge, attacking methods can be categorized into white-box and black-box attacks. \textit{White-Box attacks} refer to a setting in which the adversary can utilize all the information in the target model, including its architecture, parameters, gradient information, etc. Generally, the attacking process can be formulated as an optimization problem \cite{goodfellow2014explaining, carlini2017evaluating}. With access to such white-box information, this problem is often much easier to solve with gradient based methods. White-box attacks have been extensively studied because the disclosure of model architecture and parameters helps people understand the weakness of ML models clearly; thus, they can be analyzed mathematically. With \textit{black-Box attacks}, no knowledge of ML models is available to the adversaries. Adversaries can only feed the input data and query the outputs of the models. One of the most common ways to perform black-box attacks is to keep querying the victim model and approximate the gradient through numerical differentiation methods. Compared to white-box attacks, black-box attacks are more practical because ML models are less likely to be white-box due to privacy issues in reality.

\textbf{Targeted Attacks vs. Non-Targeted Attacks.} In the image classification problem, threat models can be categorized by whether the adversary wants to get a pre-set label for a certain image. In \textit{targeted attacks}, a specified target prediction label is expected for each adversarial example in the test phase. For example, identity theft may want to fool the face recognition system and pretend to be a specific important figure. In contrast, \textit{non-targeted attacks} expect an arbitrary prediction label except for the real one. Note that in other data domains, like graphs, the definition of targeted attack can be extended to mislead certain groups of nodes, but it is not necessary to force the model to give a particular prediction.

\subsection{Victim Models}

Victim models are the models that are attacked by the attacker. The victim model ranges from traditional machine learning models like SVM \cite{biggio2012poisoning} to Deep Neural Networks (DNNs), including the Convolution Neural Network (CNN) \cite{lecun1995convolutional}, Graph Neural Network (GNN) \cite{scarselli2008graph}, Recurrent Neural Network (RNN) \cite{zaremba2014recurrent}, etc. In this next section, we briefly intro duce the victim models that have been studied and shown to be vulnerable to adversarial attacks.

\textbf{Traditional machine learning models.} One of the earliest robustness related works checked the security of Naive Bayes classifiers \cite{dalvi2004adversarial}. Later, SVM and the naive fully-connected neural networks have been shown to be vulnerable to attacks \cite{biggio2013evasion}. Recently, the adversarial robustness of tree-based models has also been proposed as an open problem \cite{chen2019robust}.

\textbf{Deep learning models.} In computer vision tasks, Convolution Neural Networks (CNNs) \cite{krizhevsky2012imagenet} are one of the most widely used models for image classification problems. CNN models aggregate local features from images to learn the representations of image objects and give prediction based on learned representations. The vulnerability of deep neural networks to attack was first proposed in CNN \cite{szegedy2013intriguing} and since then, there has been extensive work indicating that CNNs are not robust against adversarial attacks. Graph Neural Networks (GNNs) have been developed for graph-structured data and can be used by many real-world systems, such as social networks and natural science. There are works that test GNN’s robustness \cite{Z_gner_2018, chen2018fast, dai2018adversarial, bojchevski2019adversarial, ma2019attacking} and work to build robust GNNs \cite{jin2020adversarial}. Consider the node classification problem as an example; existing works show that the performance can be reduced significantly by slightly modifying node features, adding or deleting edges, or adding fake nodes \cite{zugner2018adversarial}; for text data, one also need to consider semantic or phonetic similarity. Therefore, in addition to the commonly used optimization method used to attack a seq2seq translation model \cite{cheng2018seq2sick}, some heuristic approaches are proposed to find substitute words that attack RNN-based dialogue generation models \cite{niu2018adversarial}.


\textbf{Defense Strategies.} Under adversarial attacks, there are different types of countermeasures to prevent the adversary from creating harmful effects. During the training stage, \textit{adversarial training} aims to train a robust model by using adversarial samples during the training process. \textit{Certified defense} works to achieve robustness over all perturbations within a certain bound. For defenses that happen at inference time, \textit{adversarial example detection} works to distinguish adversarial examples, so users can reject the prediction of harmful examples.

\subsection{Representative Attack Methods}
In this subsection, we introduce representative attack methods from two aspects: evasion attacks and poisoning attacks.

\subsubsection{Evasion Attack}
Evasion attacks occur at test time. When attackers want to generate an adversarial example for a test example, they need to find a distance matrix to guarantee the perturbation size is small, so evasion attack can be further categorized by how to constrain the perturbation, i.e., pixel constrained adversarial examples with a fixed $l_p$ norm bound constraint and adversarial examples under other types of constraints.

\textbf{$L_p$ bound Attacks.} To guarantee perceptual similarity of the adversarial example and the natural example, the perturbation is normally constrained within an $l_p$ norm bound around the natural example. To find such perturbation, Projected Gradient Descent (PGD) adversarial attack \cite{madry2017towards} tries to calculate the adversarial example $x'$ that maximizes the loss function:

\begin{equation*}
    \begin{split}
        &\text{maximize  } \mathcal{L}(\theta,x')\\
        &\text{subject to  } ||x'-x||_{p}\leq\epsilon \text{  and  } x' \in [0,1]^m
    \end{split}
\end{equation*}

Here, $\epsilon$ is the perturbation budget and $m$ is the dimension of the input sample $x$. This local maximum is calculated by doing gradient ascent. At each time step, a small gradient step is made toward the direction of increasing loss value that is projected back to the $L_p$ norm bound. Currently, a representative and popular attack method is Autoattack \cite{croce2020reliable}, which is a strong evasion attack conducted by assembling four attacks, including three white-box attacks and one black-box attack, bringing a more reliable evaluation for adversarial robustness.

There are also some works with special goals. The work \cite{moosavi2017universal} devises an algorithm that successfully misleads a classifier’s decision on almost all test images. It tries to find a perturbation $\delta$ under a $\epsilon$ constraint satisfying for any sample $x$ from the test distribution. Such a perturbation $\delta$ can let the classifier give wrong decisions on most of the samples.

\textbf{Beyond $l_p$ bound attacks.}
Recently people started to realize that $l_p$ norm perturbation measurement is neither sufficient to cover real-world noise nor a perfect measurement for perceptual similarity. Some studies seek to find the minimal perturbation necessary to change the class of a given input with respect to the $l_p$ norm \cite{croce2020minimally}. Other works propose different perturbation measurements, e.g., the Wasserstein distance \cite{wu2020stronger, wong2019wasserstein}, to measure the changes of pixels.

\subsubsection{Poisoning Attacks}
As we introduced, poisoning attacks allow adversaries to take control of the training process.

\textbf{Training Time Attack.} In the training time attack, perturbation only happens in the training time. For example, the ``poisoning frog'' attack inserts an adversarial image with the true label to the training set, to make the trained model wrongly classify target test
samples \cite{shafahi2018poison}. It generates the adversarial example $x'$ by solving the following problem:

\begin{equation*}
    x' = \text{argmin}_x \|Z(x)-Z(x_t)\|_2^2+\beta \|x-x_b\|_2^2. 
\end{equation*}
Here, $Z(x)$ is the logits of the model for samples $x$, $x_t$ and $x_b$ are the samples from the target class and the original class, respectively. The result $x'$ would be similar to the base class in the input space, while sharing similar predictions with the target class. As a concrete example, some cat training samples are intentionally added as some features of bird and still labeled as “cat in training.” As a consequence, it would mislead the model’s prediction on other bird images that also contains bird features.

\textbf{Backdoor Attack.} The backdoor attack that requires perturbation happens in both training and test data. A backdoor trigger only known by the attacker is inserted into the training data to mislead the classifier into giving a target prediction on all the test examples that contain the same trigger \cite{chen2017targeted}. This type of attack is particularly dangerous because the model behaves normally on natural samples, which makes it harder to notice.

\subsection{Representative Defense Methods}
In this subsection, we introduce representative defense methods from the aforementioned categories.

\subsubsection{Robust Optimization / Adversarial Training}
Adversarial training aims to train models that give resistant predictions to adversarial examples. The training objective is formulated as a min-max problem that tries to minimize the error risk on the maximum adversarial loss within a small area around the training data samples \cite{wang2020towards}. With this bi-level optimization process, the model achieves partial robustness but still suffers from longer training time, natural and robust trade-offs, and robust overfitting issues. There are several projects making efforts to improve standard adversarial training from different perspectives. In \cite{tsipras2018robustness}, the trade-off issue is revealed. TRADES \cite{zhang2019theoretically} takes a step toward balancing the natural accuracy and robust accuracy by adding a regularization term to minimize the prediction difference on adversarial samples and natural samples. Other works \cite{wong2020fast, shafahi2019adversarial} boost the training speed by estimating the gradient of the loss for the last few layers as a constant when generating adversarial samples. They can shorten the training time to one-fourth of GPU time – or even shorter with comparable robust performance. To mitigate robust overfitting, different classic techniques, such as early stop, weight decay, and data augmentations have been investigated \cite{wu2020adversarial, rice2020overfitting}. It is evident from recent work \cite{carmon2019unlabeled} that using data augmentation methods is a promising direction to further boost adversarial training performance.

\subsubsection{Certified Defense}
Certified defense seeks to learn provably robust DNNs against specific norm-bounded perturbations \cite{raghunathan2018certified, wong2018provable}. In empirical defenses, robustness is achieved to a certain extent; in certified robust verification, however, we want to exactly answer the question whether we can find an adversarial example for a given example and the perturbation bound. For instance, a randomized smoothing based classifier \cite{cohen2019certified} aims to build an exact smooth classifier by making decisions according to the majority of predictions of neighborhood examples. Reaching such smoothness requires considerably greater computation resources, which is a challenge in practice.

\subsubsection{Detection}
In order to distinguish the adversarial examples in data distribution and to prevent the harmful effect, people often design detection algorithms. A common way to do this is to build another classifier to predict whether a sample is adversarial or not. The work \cite{gong2017adversarial} trains a binary classification model to discriminate all adversarial examples apart from natural samples and then builds ML models on recognized natural samples. Other works detect the adversarial samples based on the statistic property of adversarial sample distribution difference with natural sample distribution. In \citep{grosse2017statistical}, it uses a statistical test, i.e., a Maximum Mean Discrepancy (MMD) test to determine whether two datasets are drawn from the same distribution. It uses this tool to test whether a group of data points are natural or adversarial; however, it is shown in \cite{carlini2017adversarial} that evasion adversarial examples are not easily detected. This paper bypassing several detection methods and finds out that those defenses lack thorough security evaluations and there is no clear evidence to support that adversarial samples are intrinsically different from clean samples. Recent work proposes a new direction for black-box adversarial detection \cite{chen2020stateful}. It detects the attacker’s purpose based on the historical queries and sets a threshold for the distance between two input image queries to detect suspicious attempts to generate adversarial examples.
    

\subsection{Applications in Real Systems}
When deep learning is applied to real-world, safety-critical tasks, the existence of adversarial examples becomes more dangerous and may cause severe consequences. In the following section, we illustrate the potential threats from adversarial examples to real-world applications from different domains.

\subsubsection{Image Domain}
\begin{figure}[h]
\centering
\centering
{\includegraphics[width=0.5\linewidth]{{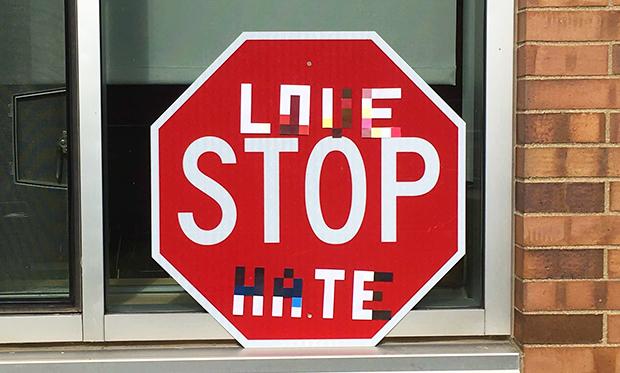}}}
\caption{The stop sign could not be distinguished by machines with modifications.}
\label{fig:roadsigh}
\end{figure}

In the auto-driving domain, road sign detection is an important task. However, with some small modifications, the road sign detection system \cite{eykholt2017robust, sitawarin2018darts} in the vehicle would recognize 35 mph as 85 mph and cannot successfully detect a stop sign as shown in Figure \ref{fig:roadsigh}. Deep learning is also widely applied in authentication tasks. An attacker can wear a special glass to pretend to be an authorized identity in order to mislead the face recognition model; this deceit can be accomplished if a few face samples labeling the glasses as the target identity are inserted into the training set \cite{chen2017targeted}. Person detection can also be avoided through the wearing of an adversarial T-shirt \cite{xu2020adversarial}.

\subsubsection{Text Domain}
Adversarial attacks also happen in many natural language processing tasks, including text classification, machine translation, and dialogue generation. For machine translation, sentence and word paraphrasing on input texts are conducted to craft adversarial examples \citep{DBLP:journals/corr/abs-1812-00151}. The attack method first builds a paraphrasing corpus that contains a lot of words and sentence paraphrases. To find an optimal paraphrase of an input text, a greedy method is adopted to search valid paraphrases for each word or sentence from the corpus. Moreover, it proposes a gradient-guided method to improve the efficiency of the greedy search. In \citep{liu2019say}, it treats the neural dialogue model as a black-box and adopts a reinforcement learning framework to effectively find trigger inputs for targeted responses. The black-box setting is stricter but more realistic, while the requirements for the generated responses are properly relaxed. The generated responses are expected to be semantically identical to the targeted ones but not necessarily exactly match with them.

\subsubsection{Audio Data}

The state-of-art speech-to-text transcription networks, such as DeepSpeech \cite{hannun2014deep}, can be attacked by a small perturbation \cite{carlini2018audio}. Given any speech waveform $x$, an inaudible sound perturbation $\delta$ is added to make the synthesized speech $x+\delta$ recognized as any targeted desired phrase. In the work \cite{saadatpanah2020adversarial}, it proposes an adversarial attack method toward the YouTube CopyRight detection system to avoid music with copyright issues being detected during uploading. The attack uses a neural network to extract features from the music piece to create a fingerprint, which is used for checking whether the music matches existing copyrighted music. By doing a gradient-based adversarial attack on the original audio piece to create a large difference on the output fingerprint, the modified audio can successfully avoid the detection of YouTube's CopyRight detection system.

\subsubsection{Graph Data}
\citet{zugner2018adversarial} consider attacking node classification models, graph convolutional networks \cite{kipf2016semi}, by modifying the node connections or node features. In this setting, an adversary is allowed to add or remove edges between nodes or change the node features with a limited number of operations in order to mislead the GCN model that is trained on the perturbed graph. The work \cite{zugner2019adversarial} attempts to poison the graph so that the global node classification performance of GCN will drop – and even be made almost useless. They optimize the graph structure as the hyper-parameters of the GCN model with the meta-learning technique. The goal of the attack \cite{bojcheski2018adversarial} is to perturb the graph structure to corrupt the quality of node embedding, affecting the downstream tasks’ performance, including node classification or link prediction. Pro-GNN \cite{jin2020graph} tries to jointly learn a structural graph and a graph neural network from the perturbed graph guided by some intrinsic properties of a real-world graph, such as low-rank and feature smoothness. The defense method could learn a clean adjacency matrix close to the perturbed one and constrain its norm to guarantee the low-rank property. Meanwhile, a feature smoothness regularizer is also utilized to penalize rapid changes in features between adjacent nodes. A robust GNN would build upon the learned graph.

\subsection{Surveys and Tools}
\label{sec:rob_repo}
In this subsection, we list current surveys and tools on adversarial robustness to help readers get easy access to resources.

\subsubsection{Surveys}
\citet{xu2020adversarial} give a comprehensive introduction of concepts and go through representative attack and defense algo-rithms in different domains, including image classification, graph classification, and natural language processing. For the surveys in a specific domain, \citet{akhtar2018threat} provide a comprehensive introduction to adversarial threats in a computer vision domain \cite{chakraborty2018adversarial}; \citet{jin2020adversarial} give a thorough review on the latest adversarial robustness techniques in the graph domain; and \citet{zhang2020adversarial} focus on natural language processing and summarize important algorithms on adversarial robustness in the text domain.

\subsubsection{Tools}
\textit{Advertorch} \cite{ding2019advertorch} is a Pytorch toolbox containing many popular attack methods in the image domain. \textit{DeepRobust} \cite{li2020deeprobust} is a comprehensive and up-to-date adversarial attacks and defenses library based on Pytorch that includes not only algorithms in the image domain but also the graph domain. This platform provides convenient access to different algorithms and evaluation functions to illustrate the robustness of image classification models or graph properties. \textit{RobustBench} \cite{croce2021robustbench} provides a robust evaluation platform by the Autoattack algorithm for different adversarial training models. This platform also provides well-trained robust models by different adversarial training methods, which can save resources for researchers.

\subsection{Future Directions}
\label{sec:rob_fut}
For adversarial attacks, people are seeking more general attacking methods to evaluate adversarial robustness. For black-box attacks, efficiently generating adversarial examples with fewer adversarial queries is often challenging. For training in adversarial defense methods, an important issue is the robust overfitting and lack of generalization in both adversarial and natural examples. This problem remains unsolved and needs further improvement. Another direction for adversarial training is to build robust models against more general adversarial examples, including but not limited to different l-p bound attacks. For certified defenses, one possible direction is to train a model with robust guarantee more efficiently, since the current certified defense methods require a large number of computational resources.
\section{Non-discrimination \& Fairness}
\label{sec:fairness}
A trustworthy AI system ought to avoid discriminatory behaviors in human-machine interactions and ensure fairness in decision making for any individuals or groups. With the rapid spread of AI systems in our daily lives, more and more evidence indicates that AI systems show human-like discriminatory bias or make unfair decisions. For example, Tay, the online AI chatbot developed by Microsoft, produced a lot of improper racist and sexist comments, which led to its closure within 24 hours of release \cite{wolf2017we}; dialogue models trained on human conversations show bias toward females and African Americans by generating more offensive and negative responses for these groups \cite{liu2020does}. Moreover, a recidivism prediction software used by U.S. courts often assigns a higher risky score for an African American than a Caucasian with a similar profile\footnote{\url{https://www.propublica.org/article/machine-bias-risk-assessments-in-criminal-sentencing}} \cite{mehrabi2019survey}; a job recommendation system promotes more STEM employment opportunities to male candidates than to females \cite{lambrecht2019algorithmic}. As AI plays an increasingly irreplaceable role in promoting the automation of our lives, fairness in AI is closely related to our vital interests and demands considerable attention. Recently, many works have emerged to define, recognize, measure, and mitigate the bias in AI algorithms. In this section, we aim to give a comprehensive overview of the cutting-edge research progress addressing fairness issues in AI. In the subsections, we first present concepts and definitions regarding fairness in AI.We then provide a detailed taxonomy to discuss different origins of algorithmic bias, different types of bias, and fairness.We then review and classify popular bias mitigation technologies for building fair AI systems. Next, we introduce the specific bias issues and the applications of bias mitigation methods in real-world AI systems. In this part, we categorize the works according to the types of data processed by the system. Finally, we discuss the current challenges and future opportunities in this field. We expect that researchers and practitioners can gain a sense of direction and understanding from a broad overview of bias and fairness issues in AI and deep insight into the existing solutions, so as to advance progress in this field.

\subsection{Concepts and Taxonomy}
Before we go deep into nondiscrimination and fairness in AI, we need to understand how relative concepts, such as bias and fairness, are defined in this context. In this subsection, we briefly illustrate the concepts of bias and fairness, and provide a taxonomy to introduce different sources of bias, different types of bias, and fairness.

\subsubsection{Bias}
In the machine learning field, the word ``bias'' has been abused. It conveys different meanings in different contexts. We first distinguish the concept of bias in the context of AI non-discrimination and fairness from that in other contexts. There are three categories of bias: \textbf{productive bias}, \textbf{erroneous bias}, and \textbf{discriminatory bias}. Productive bias exists in all machine learning algorithms. It is beneficial and necessary for an algorithm to be able to model the data and make decisions \cite{hildebrandt2019privacy}. Based on the ``no free lunch theory'' \cite{wolpert1997no}, only a predictive model biased toward certain distributions or functions can achieve better performance on modeling them. Productive bias helps an algorithm to solve certain types of problems. It is introduced through our assumptions about the problem, which is specifically reflected as the choice of a loss function, an assumed distribution, or an optimization method, etc. Erroneous bias can be viewed as a systematic error caused by faulty assumptions. For example, we typically assume that the distribution of the training data is consistent with the real data distribution. However, due to selection bias \cite{marlin2012collaborative} or sampling bias \cite{mehrabi2019survey}, the collected training data may not be able to represent the real data distribution. Thus, the violation of our assumption can lead to the learned model’s undesirable performance on the test data. Discriminatory bias is the kind of bias we are interested in under AI nondiscrimination and fairness. As opposed to fairness, discriminatory bias reflects an algorithm’s unfair behaviors toward a certain group or individual, such as producing discriminatory content or performing less well for some people \cite{shah2020predictive}. In the rest of this paper, when we mention bias, we refer to discriminatory bias.

\textbf{Sources of Bias.} The bias in an AI system can be produced by different sources, namely, the data, the algorithm, or the evaluation method. Bias within data comes from different phases of data generation, from data annotation to data collection to data processing \cite{olteanu2019social,shah2019predictive}. In the phase of data annotation, bias can be introduced due to a non-representative group of annotators \cite{joseph2017constance}, inexperienced annotators \cite{plank2014learning}, or preconceived stereotypes held by the annotators \cite{sap2019risk}. In the phase of data collection, bias can emerge due to the selection of data sources or how data from several different sources are acquired and prepared \cite{olteanu2019social}. In the data processing stage, bias can be generated due to data cleaning \cite{denny2016assessing}, data enrichment \cite{cohen2013classifying}, and data aggregation \cite{tufekci2014big}.

\textbf{Types of Bias.} Bias can be categorized into different classes from different perspectives. It can be explicit or implicit. \textit{Explicit bias}, also known as direct bias, occurs when the sensitive attribute explicitly causes an undesirable outcome for an individual; while \textit{implicit bias}, also known as indirect bias, indicates the phenomenon that an undesirable outcome is caused by nonsensitive and seemingly neutral attributes, which in fact have some potential associations with the sensitive attributes \cite{Zhang2017ACF}. For example, the residential address seems a nonsensitive attribute, but it can correlate with the race of a person according to the population distribution of different ethnic groups \cite{Zhang2017ACF}. Moreover, language style can reflect the demographic features of a person, such as race and age \cite{huang2020multilingual,liu2021authors}. Bias can be acceptable and unacceptable.\textit{Acceptable bias}, also known as explainable bias, describes a situation where the discrepancy of outcomes for different individuals or groups can be reasonably explained by factors. For example, models trained on the UCI Adult dataset predict higher salaries for males than females. Actually, this is because males work for a longer time per week than females \cite{Kamiran2013ExplainableAN}. Based on this fact, such biased outcomes are acceptable and reasonable. Conversely, bias that cannot be explained appropriately is treated as \textit{unacceptable bias}, which should be avoided in practice.

\subsubsection{Fairness}
The fairness of an algorithm is defined as ``the absence of any prejudice or favoritism towards an individual or a group based on their intrinsic or acquired traits in the context of decision making'' \cite{saxena2019fairness,mehrabi2019survey}. Furthermore, according to the object of the study, fairness can be further defined as group fairness and individual fairness.

\textbf{Group Fairness.} Group fairness requires that two groups of people with different sensitive attributes receive comparable treatments and outcomes statistically. Based on this principle, various definitions have been proposed, such as Equal Opportunity \cite{Hardt2016EqualityOO}, which requires people from two groups to be equally likely to get a positive outcome when they indeed belong to the positive class; Equal Odds \cite{Hardt2016EqualityOO}, which requires that the probability of being classified correctly should be the same for different groups; and Demographic Parity \cite{dwork2012fairness}, which requires different groups to have the same chance to get a positive outcome, etc.

\textbf{Individual Fairness.} While group fairness can maintain fair outcomes for a group of people, a model can still behave discriminatorily at the individual level \cite{dwork2012fairness}. Individual fairness is based on the understanding that similar individuals should be treated similarly. A model satisfies individual fairness if it gives similar predictions to similar individuals \cite{dwork2012fairness,Kusner2017CounterfactualF}. Formally, if individuals $i$ and $j$ are similar under a certain metric $\delta$, the difference between the predictions given by an algorithm $M$ on them should be small enough: $|f_M(i)-f_M(j)|<\epsilon$, where $f_M(\cdot)$ is the predictive function of algorithm $M$ that maps an individual to an outcome, and $\epsilon$ is a small constant.

\subsection{Methods}
In this subsection, we introduce bias mitigation techniques. Based on which stage of an AI pipeline is to interfere, the debiasing methods can be categorized into three types: \textbf{pre-processing}, \textbf{in-processing} and \textbf{post-processing} methods. Representative bias mitigation methods are summarized in Table \ref{table:debias}.

\begin{table}[ht]
	\centering
	\caption{Representative debiasing strategies in the three categories.}
	\label{table:debias}
	\begin{tabular}{@{}|c|c|c|@{}}
        \hline
		\textbf{Category} & \textbf{Strategy} & \textbf{References} \\
		\hline
		\multirow{5}{*}{Pre-processing} & Sampling & \cite{zhang2016identifying,adler2018auditing,bastani2019probabilistic} \\
		&  Reweighting & \cite{calders2013unbiased,kamiran2012data,zhang2020demographics} \\
		&  Blinding & \cite{hardt2016equality,chen2018my,chouldechova2017fairer,zafar2017fairness} \\
		&  Relabelling & \cite{hajian2012methodology,kamiran2012data,cowgill2017algorithmic} \\
		&  Adversarial Learning & \cite{adel2019one,feng2019learning,kairouz2019censored} \\
		\hline
		\multirow{4}{*}{In-processing} & Reweighting & \cite{krasanakis2018adaptive,jiang2020identifying} \\
		&  Regularization & \cite{feldman2015certifying,aghaei2019learning} \\
		&  Bandits & \cite{liu2017calibrated,ensign2018decision} \\
		&  Adversarial Learning & \cite{zhang2018mitigating,celis2019improved,liu2020mitigating,liu2021authors} \\
		\hline
		\multirow{3}{*}{Post-processing} & Thresholding & \cite{Hardt2016EqualityOO,Menon2017TheCO,Iosifidis2019FAEAF} \\
		&  Transformation & \cite{Kilbertus2017AvoidingDT,Nabi2018FairIO,Chiappa2019PathSpecificCF} \\
		&  Calibration & \cite{HbertJohnson2017CalibrationFT,Kim2018FairnessTC} \\
    \hline
	\end{tabular}
	\vspace{-3mm}
\end{table}

\textbf{Pre-processing Methods.} Pre-processing approaches try to remove the bias in the training data to ensure the fairness of an algorithm from the origin \cite{kamiran2012data}. This category of methods can be adopted only when we have access to the training data. Various strategies are proposed to interfere with training data. Specifically, \citet{Celis2016HowTB} propose to adaptively sample the instances that are both diverse in features and fair to sensitive training attributes. Moreover, reweighting methods \cite{Kamiran2011DataPT,zhang2020demographics} try to mitigate the bias in training data by adaptively up-weighting the training instances of underrepresented groups, while down-weighting those of overrepresented groups. Blinding methods try to make a classifier insensitive to a protected variable. For example, \citet{hardt2016equality} force a classifier to have the same threshold value for different race groups to ensure that the predicted loan rate is equal for all races. Some works \cite{Kamiran2011DataPT,Zemel2013LearningFR} try to relabel the training data to ensure the proportion of positive instances are equal across all protected groups. Additionally, \citet{Xu2018FairGANFG} take advantage of a generative adversarial network to produce bias-free and high-utility training data.

\textbf{In-processing Methods.} In-processing approaches address the bias at the algorithm level and try to eliminate bias during the model training process. They often seek to create a balance between performance and fairness \cite{caton2020fairness}. \citet{krasanakis2018adaptive} propose an in-processing re-weighting approach. They first train a vanilla classifier to learn the weights of samples and then retrain the classifier using these weights. Some works \cite{Kamishima2012FairnessAwareCW,Goel2018NonDiscriminatoryML} take advantage of regularization methods, where one or more penalty terms are added into the objective function to penalize biased outcomes. The idea of adversarial learning is also adopted in in-processing debiasing methods. \citet{liu2020mitigating} design an adversarial learning framework to train neural dialogue models that are free from gender bias. Alternatively, bandits recently emerged as a novel idea for solving fairness problems. For example, \citet{Joseph2016FairAF} propose solving the fairness problem under a stochastic multi-armed bandit framework with fairness metrics as the rewards and the individuals or groups under investigation as the arms.

\textbf{Post-processing Methods.} Post-processing approaches directly make transformations on the model’s outputs to ensure fair final outcomes. \citet{hardt2016equality} propose approaches to determine threshold values via measures such as equalized odds specifically for different protected groups to find a balance between the true and false positive rates to minimize the expected classifier loss. \citet{Feldman2015CertifyingAR} propose a transformation method to learn a new fair representation of the data. Specifically, they transform the SAT score into a distribution of the rank order of the students independent of gender. \citet{Pleiss2017OnFA} borrow the idea of calibration to build fair classifiers. Similar to the traditional definition of calibration that the proportion of positive predictions should be equal to the proportion of positive examples, they force the conditions to hold for different groups of people. Nevertheless, they also find that there is a tension between prediction accuracy and calibration.

\subsection{Applications in Real Systems}
In this subsection, we summarize the studies regarding bias and fairness issues in real-world AI systems during different tasks. We introduce the works following the order of different data domains, including tabular data, images, texts, audios, and graphs. For each domain, we describe several representative tasks and present how AI systems can be biased on these tasks. A summary of the representative works can be found in Table \ref{table:bias_detection}.

\begin{table*}[ht]
	\centering
	\caption{A summary of bias detection works in different data domains.}
	\label{table:bias_detection}
	\begin{tabular}{@{}|c|c|l|@{}}
		\hline
		\textbf{Domain} & \textbf{Task} & \textbf{References} \\
		\hline
		\multirow{3}{*}{Tabular Data} & Classification & \cite{kamiran2009classifying,calders2009building,calders2010three,hardt2016equality,goel2018non,menon2018cost}\\
		\cline{2-3}
		&  Regression & \cite{berk2017convex,agarwal2019fair}\\
		\cline{2-3}
		&  Clustering & \cite{backurs2019scalable,chen2019proportionally}\\ \hline
		\multirow{3}{*}{Image Data} & Image Classification & \cite{pachal2015google} \\
		\cline{2-3}
		&  Face Recognition & \cite{buolamwini2018gender,howard2018ugly}\\
		\cline{2-3}
		&  Object Detection & \cite{ryu2017improving}\\
		\hline
		\multirow{5}{*}{Text Data} & Text Classification & \cite{kiritchenko2018examining,park2018reducing,dixon2018measuring,borkan2019nuanced,zhang2020demographics,huang2020multilingual}\\
		\cline{2-3}
		& Embedding & \cite{bolukbasi2016man,brunet2019understanding,gonen2019lipstick,zhao2019gender,papakyriakopoulos2020bias,may2019measuring}\\
		\cline{2-3}
		&  Language Modeling & \cite{bordia2019identifying,sheng2019woman,lu2020gender,gehman2020realtoxicityprompts,yeo2020defining}\\ 
		\cline{2-3}
		&  Machine Translation & \cite{vanmassenhove2019getting,stanovsky2019evaluating,cho2019measuring,basta2020towards,gonen2020automatically}\\ 
        \cline{2-3}
		&  Dialogue Generation & \cite{liu2020does,dinan2020queens,curry2020conversational}\\
		\hline
		Audio Data & Speech Recognition & \cite{rodger2004field,carty2011many,tatman2016google,howard2018ugly}\\
		\hline
		\multirow{2}{*}{Graph Data} & Node Embedding & \cite{bose2019compositional} \\
		\cline{2-3}
		&  Graph Modeling & \cite{dai2021say} \\ 
    \hline
	\end{tabular}
\end{table*}

\subsubsection{Tabular Domain}
Tabular data is the most common format of data in machine learning; thus, the research on bias in machine learning is predominantly conducted on tabular data. In the recent decade, researchers have investigated how algorithms can be biased in classification, regression, and clustering tasks. For classification, researchers find evidence that machine learning models for credit prediction \cite{kamiran2009classifying} and recidivism prediction \cite{Chouldechova2017FairPW} tasks can show significant prejudice toward certain demographic attributes of a person, such as race and gender. \citet{berk2017convex} and \citet{agarwal2019fair} investigate multiple regression tasks, from salary estimation to crime rate prediction, showing unfair treatment for different races and genders. \cite{backurs2019scalable} and \cite{chen2019proportionally} evaluate the fairness in clustering algorithms with a belief that as data points, different groups of people are entitled to be clustered with the same accuracy.

\subsubsection{Image Domain}
Machine learning models in computer vision have also shown unfair behaviors. In \cite{buolamwini2018gender,howard2018ugly}, the authors showed that face recognition systems work better for white compared to darker faces. An image classification application developed by Google has been accused of labeling black people as ``gorillas'' \cite{pachal2015google}. The work in \cite{howard2018ugly} balanced the dataset for face recognition tasks aimed at alleviating gender bias. In \cite{ryu2017improving}, the authors employed a transfer learning method and improved smiling detection against gender and race discrimination. The work \cite{zhao2017men} tackled the social bias in visual semantic role labeling, e.g., associating cooking roles with women. They introduced corpus-level constraints for calibrating existing structured prediction models. In work \cite{wang2020towards}, a visual recognition benchmark is designed for studying bias mitigation.

\subsubsection{Text Domain}
A large number of works have shown that algorithmic bias exists in various natural language processing tasks. Word embeddings often exhibit a stereotypical bias for text data, causing a serious risk of perpetuating problematic biases in imperative societal contexts. In \cite{NIPS2016_6228}, the authors first showed that popular state-of-the-art word embeddings regularly mapped men to working roles and women to traditional gender roles, leading to significant gender bias and even downstream tasks. Following the research of word embeddings, the same patterns of gender bias are discovered in sentence embeddings \cite{DBLP:journals/corr/abs-1903-10561}. In the task of co-reference resolution, researchers demonstrated in \cite{DBLP:journals/corr/abs-1804-06876} that rule-based, feature-based, and neural network-based co-reference systems all show gender bias by linking gendered pronouns to pro-stereotypical entities with higher accuracy than anti-stereotypical entities. Language models can also learn gender discrimination from man-made text data \cite{DBLP:journals/corr/abs-1904-03035}, which tend to generate certain words reflecting gender stereotypes with different probabilities in the context of males and females. As for machine translation, it has been illustrated that Google’s translation system suffers from gender bias by showing favoritism toward males for stereotypical fields, such as STEM jobs when translating sentences taken from the U.S. Bureau of Labor Statistics into a dozen gender-neutral languages \cite{DBLP:journals/corr/abs-1809-02208}. Dialogue systems, including generative models and retrieval- based models, also show bias toward different genders and races by producing discriminatory responses \cite{liu2020does,liu2020mitigating}.

\subsubsection{Audio Domain}
Voice recognition systems show gender bias by processing the voices of men and women differently \cite{howard2018ugly}. It was found that medical voice-dictation systems recognize voice inputs from males versus females with higher accuracy \cite{rodger2004field}. It was shown in \cite{carty2011many} that voice control systems on vehicles worked better for males than females. Google’s speech recognition software can understand queries from male voices more consistently than those from females \cite{tatman2016google}.

\subsubsection{Graph Domain}
ML applications on graph-structured data are ubiquitous in the real world. The fairness issues in these problems are drawing increasing attention from researchers. Existing graph embedding techniques can learn node representations correlated with protected attributes, such as age and gender. Consequently, they exhibit bias toward certain groups in real-world applications, like social network analysis and recommendations \cite{bose2019compositional}. Graph neural networks (GNNs) also inherit bias from training data and even magnify the bias through GNN’s graph structures and message-passing mechanism \cite{dai2021say}.

\subsection{Surveys and Tools}
In this subsection, we gather the existing surveys, tools and repositories on fairness in AI to facilitate readers wishing to explore this field further.

\subsubsection{Surveys}
The problem of fairness has been studied in multiple disciplines other than computer science for more than a half century. In one survey \cite{hutchinson201950}, the authors trace the evolution of the notions and measurements of fairness in different fields, such as education and hiring, over the past 50 years. They provide a comprehensive comparison between the past and current definitions to encourage a deeper understanding of modern fairness in AI. \citet{zliobaite2015survey} provides an early survey on measuring indirect discrimination in machine learning. In this survey, the authors review early approaches for measuring bias in data and predictive models. They also analyze the measurements from other fields and explore the possibility of their use in machine learning. \citet{corbett2018measure} provide a critical review on the measurements of fairness, showing the limitations of the existing fairness criteria in classification tasks in machine learning. \citet{mehrabi2019survey} contribute a comprehensive survey on bias and fairness in machine learning. In this survey, the authors provide a detailed taxonomy of the bias and fairness definitions in machine learning, and also introduce the bias observed in the data and algorithms in different domains of AI and the state-of-the-art debiasing methods. \citet{caton2020fairness} provide an overview of the existing debiasing approaches for building fair machine learning models. They organize the extant works into three categories and 11 method areas and introduce them following their taxonomy. Moreover, there are some surveys regarding bias and fairness in specific domains of AI. \citet{blodgett2020language} review the papers analyzing bias in NLP systems, providing critical comments on such works and indicating that many existing works suffer from unclear and inconsistent motivations and irrational reasoning. They also offer suggestions to normalize future studies on bias in NLP. \citet{chen2020bias} summarize and organize the works on bias and debias in recommender systems, and discuss future directions in this field.

\subsubsection{Tools}
\label{sec:fair_tool}
In recent years, some organizations or individual researchers have provided multi-featured toolkits and repositories to facilitate fair AI. The repository \textit{Responsibly} \cite{louppe2016learning} collects the datasets and measurements for evaluating bias and fairness in classification and NLP tasks. The project \textit{FairTest} \cite{tramer2017fairtest} provides an unwarranted associations (UA) framework to discover unfair user treatment in data-driven algorithms. \textit{AIF360} \cite{bellamy2018ai} collects popular datasets for fairness studies and provides the implementations of common debiasing methods for binary classification. \textit{Aequitas} \cite{saleiro2018aequitas} is released as an audit toolkit to test the bias and fairness of models for multiple demographic groups on different metrics. The repository \textit{Fairness Measurements}\footnote{\url{https://github.com/megantosh/fairness_measures_code/tree/master}} provides datasets and codes for quantitatively measuring discrimination in classification and ranking tasks.

\subsection{Future Directions}
Fairness research still possesses a number of outstanding challenges.
\begin{itemize}
   \item \textbf{Trade-off between fairness and performance.} Studies on fairness in different fields have confirmed the existence of the trade-off between fairness and performance of an algorithm \cite{corbett2017algorithmic,prost2019toward,berk2021fairness}. The improvement of the fairness of an algorithm typically comes at the cost of performance degradation. Since both fairness and performance are indispensable, extensive research is needed to help people better understand an algorithm’s trade-off mechanism between them, so that practitioners can adjust the balance in practical usage based on the actual demand;
    
    \item \textbf{Precise conceptualization of fairness.} Although extensive research has been conducted on bias and fairness in AI, too much of this work formulates its concerns under a vague concept of bias that refers to any system harmful to human behaviors but fails to provide a precise definition of bias or fairness specific to their setting \cite{blodgett2020language}. In fact, different forms of bias can appear in different tasks, even in the same task. For example, in a recommender system, popularity bias can exist toward both the users and items \cite{chen2020bias}. In a toxicity detection algorithm, race bias can exist toward both the people mentioned in texts and the authors of texts \cite{liu2021authors}. To study any fairness problem, a precise definition of bias indicating how, to whom, and why an algorithm can be harmful must be articulated. In this way, we can make the research on AI fairness in the whole community more standardized and systematic;
    
    \item \textbf{From equality to equity.} Fairness definitions are often associated with equality to ensure that an individual or a conserved group, based on race or gender, are given similar amounts of resources, consideration, and results. Nonetheless, the area of equity has been heavily under-examined \cite{mehrabi2019survey}, where this notion pertains to the particular resources for an individual or a conserved group to be successful \cite{gooden2015race}. Equity remains an interesting future direction, since the exploration of this definition can extend or contradict existing definitions of fairness in machine learning.
\end{itemize}
\section{Explainability}
\label{sec:interpret}

The improved predictive performance of AI systems has often been achieved through increased model complexity \cite{doshi2017towards,molnar2020interpretable}. A prime example is the paradigm of deep learning, dominating the heart of most state-of-the-art AI systems. However, deep learning models are treated as black-boxes, since most of them are too complicated and opaque to be understood and are developed without explainability \cite{linardatos2021explainable}. More importantly, without explaining the underlying mechanisms behind the predictions, deep models cannot be fully trusted, which prevents their use in critical applications pertaining to ethics, justice, and safety, such as healthcare \cite{miotto2018deep}, autonomous cars \cite{levinson2011towards}, and so on. Therefore, building a trustworthy AI system requires understanding of how particular decisions are made \cite{world2020future}, which has led to the revival of the field of eXplainable Artificial Intelligence (XAI). In this section, we aim to provide intuitive understanding and high-level insights on the recent progress of explainable AI. First, we provide the concepts and taxonomy regarding explainability in AI. Second, we review representative explainable techniques for AI systems according to the aforementioned taxonomy. Third, we introduce real-world applications of explainable AI techniques. Finally, we provide some surveys and tools and discuss future opportunities on explainable AI.

\subsection{Concepts and Taxonomy}
In this subsection, we introduce the concepts of explainability in AI. We then provide a taxonomy of different explanation techniques.

\subsubsection{Concepts}
In the context of machine learning and AI literature, explainability and interpretability are usually used by researchers interchangeably \cite{molnar2020interpretable}. One of the most popular definitions of explainability is the one from Doshi-Velez and Kim, who define it as ``the ability to explain or to present in understandable terms to a human'' \cite{doshi2017towards}. Another popular definition is from Miller, where he defines explainability as ``the degree to which a human can understand the cause of a decision'' \cite{miller2019explanation}. In general, the higher the explainability of an AI system is, the easier it is for someone to comprehend how certain decisions or predictions have been made. Meanwhile, a model is better explainable than other models if its decisions are easier for a human to comprehend than those of others.

While explainable AI and interpretable AI are very closely related, subtle differences between them are discussed in some studies \cite{rudin2019stop,gilpin2018explaining,zhang2018explainable}.

\begin{itemize}
\item A model is interpretable if the model itself is capable of being understood by humans on its predictions. When looking at the model parameters or a model summary, humans can understand exactly the procedure on how it made a certain prediction/decision and, even given a change in input data or algorithmic parameters, it is the extent to which humans can predict what is going to happen. In other words, such models are intrinsically transparent and interpretable, rather than black-box/opaque models. Examples of interpretable models include decision trees and linear regression.

\item An explainable model indicates that additional (post hoc) explanation techniques are adopted to help humans understand why it made a certain prediction/decision that it did, although the model is still black-box and opaque. Note that such explanations are often not reliable and can be misleading. Examples of such models would be deep neural networks based models, where the models are usually too complicated for any human to comprehend.
\end{itemize}

\subsubsection{Taxonomy}
Techniques for AI’s explanation can be grouped according to various criteria.

\begin{itemize}
\item  \textbf{Model usage: model-intrinsic and model-agnostic.} If the application of interpretable techniques is only restricted to a specific architecture of an AI model, then these interpretable techniques are called a model-intrinsic explanation. In contrast, the techniques that could be applied in every possible algorithm are called a model-agnostic explanation.

\item \textbf{Scope of Explanation: local and global.} If the method provides an explanation only for a specific instance, then it is a local explanation; if the method explains the whole model, then it is a global explanation.

\item \textbf{Differences in the methodology: gradient-based and perturbation-based.} If the techniques employ the partial derivatives on input instances to generate attributions, then these techniques are called a gradient-based explanation; if the techniques focus on the changes or modifications of input data, we name them a perturbation-based explanation.

\item \textbf{Explanation via Other Approaches: Counterfactual Explanations.} We present other explanation techniques that cannot be easily categorized with the previous groups. A counterfactual explanation usually refers to a causal situation in the form, ``If $X$ had not occurred, $Y$ would not have occurred.'' In general, counterfactual explanation methods are model-agnostic and can be used to explain predictions of individual instances (local) \cite{wachter2017counterfactual}.
\end{itemize}

\subsection{Methods}

In this subsection, we introduce some representative explanation techniques according to the aforementioned taxonomy. A summary of the representative works can be found in Table~\ref{tab:XAI}.

\begin{figure}[tb]
\centering
\centering
{\includegraphics[width=0.593\linewidth]{{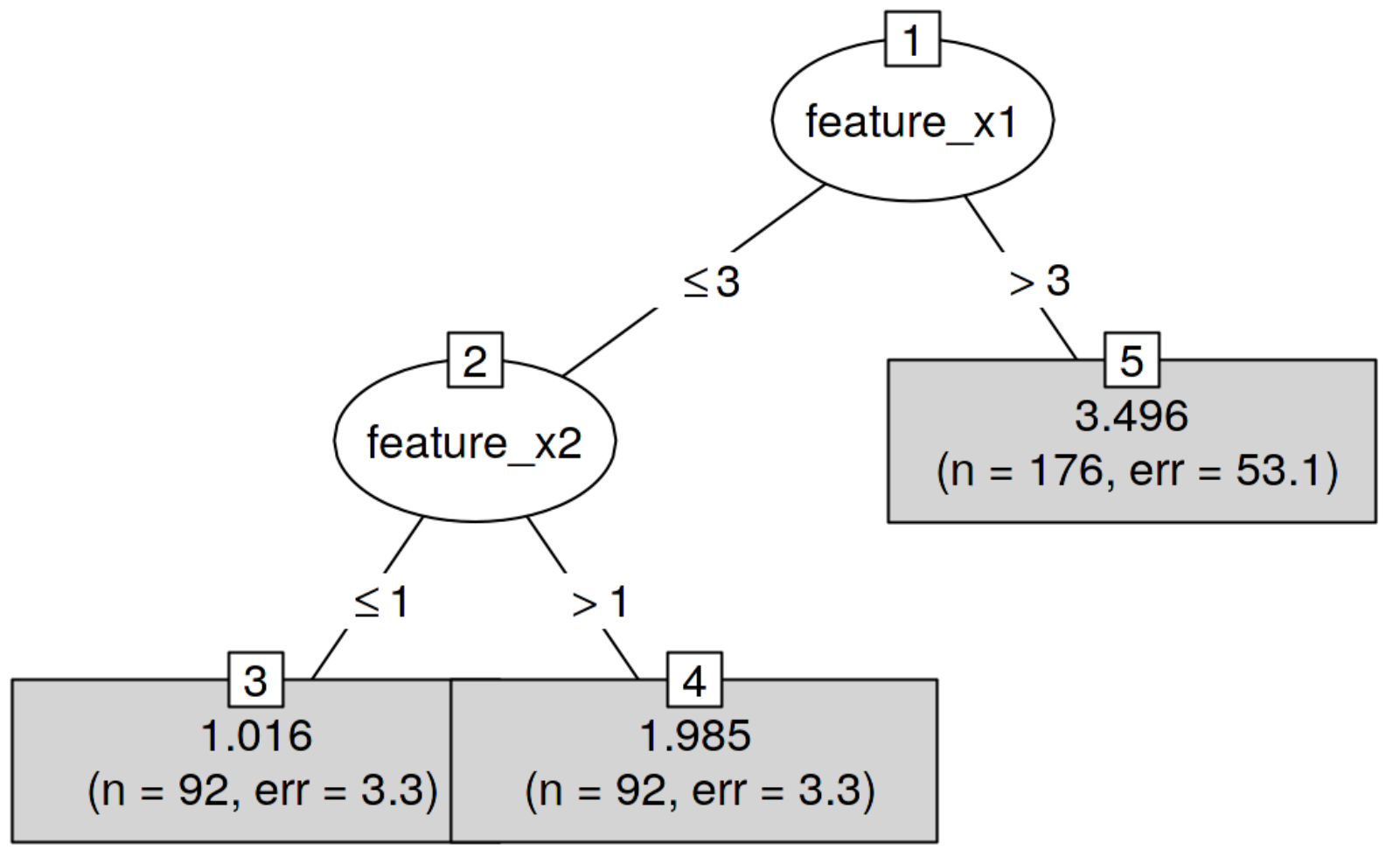}}}
\caption{  The interpretation of decision tree is simple, where intermediate nodes in the tree represent decisions and leaf nodes can be class labels.  Starting from the root node to leaf nodes can create good explanations on how the certain label is made by decision tree model. (Image Credit: \cite{molnar2020interpretable}) }
\label{fig:dt}
\end{figure}

\subsubsection{{Model usage}}

Any explainable algorithm that is dependent on the model architecture can fall into the model-intrinsic category. In contrast, model-agnostic methods apply to any model for being generally applicable. In general, there are significant research interests in developing model-agnostic methods to explain the predictions of an existing well-performing neural networks model. This criterion also can be used to distinguish whether interpretability is achieved by restricting the complexity of the AI model. Intrinsic interpretability refers to AI models that are considered interpretable (white-box) due to their simple model architecture, while most model-agnostic explanations are widely applied into (black-box) deep neural networks which are highly complicated and opaque due to their millions of parameters.

\begin{itemize}
\item \textbf{Model-intrinsic Explanations}

The model in this category is often called an intrinsic, transparent, or white-box explanation. Generally, without designing an additional explanation algorithm, this type of interpretable technique cannot be re-used by other classifier architectures. Therefore, the model intrinsic methods of explanations are inherently model specific.
Such commonly used interpretable models include linear/logistic regression,  decision trees, rule-based models, Generalized Additive Models (GAMs), Bayesian networks, etc.

For example, the linear regression model~\cite{bishop2006pattern}, as one of the most representative linear models in ML, aims to predict the target as a weighted sum of the feature of instances. With this linearity of the learned relationship, the linear regression model makes the estimation procedure simple and significantly understandable on a modular level (i.e., the weights) for humans.
Mathematically, given one instance with $d$ dimension of features $\mathbf{x}$, the linear regression model can be used to model the dependence of a  predicted target $\hat{y}$  as follows:
\begin{align}
\hat{y}=\mathbf{w}^T\mathbf{x} + b=  w_1 x_1 + ... + w_d x_d + b
\end{align}
where $\mathbf{w}$ and $b$ denote the learned feature weights and the bias term, respectively.  The predicted target $\hat{y}$ of linear regression is a weighted sum of its $d$ dimension features $\mathbf{x}$ for any instance, where the decision-making procedure is easy for a human to comprehend by inspecting the value of the learned feature weights $\mathbf{w}$. 

Another representative method is decision tree~\cite{quinlan1986induction}, which contains a set of conditional statements arranged hierarchically. Making predictions in decision tree is also the procedure of explaining the model by seeking the path from the root node to leaf nodes (label), as illustrated in Figure~\ref{fig:dt}.

\begin{figure}[tb]
\centering
\centering
{\subfigure[Original Image]
{\includegraphics[width=0.2393\linewidth]{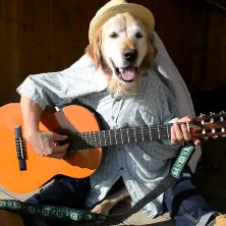}}}
{\subfigure[Electric Guitar: 0.32]
{\includegraphics[width=0.2393\linewidth]{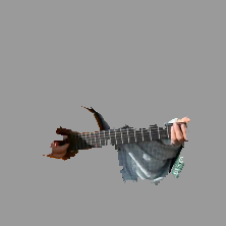}}}
{\subfigure[Acoustic Guitar: 0.24]
{\includegraphics[width=0.2393\linewidth]{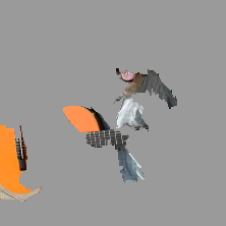}}}
{\subfigure[Labrador: 0.21]
{\includegraphics[width=0.2393\linewidth]{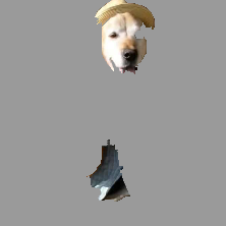}}}
\caption{  LIME explains deep neural networks on image classification task: top 3 predicted categories and the corresponding scores. (Image Credit: \cite{ribeiro2016should}) }
\label{fig:lime}
\end{figure}

\begin{figure}[tb]
\centering
\centering
{\includegraphics[width=0.63\linewidth]{{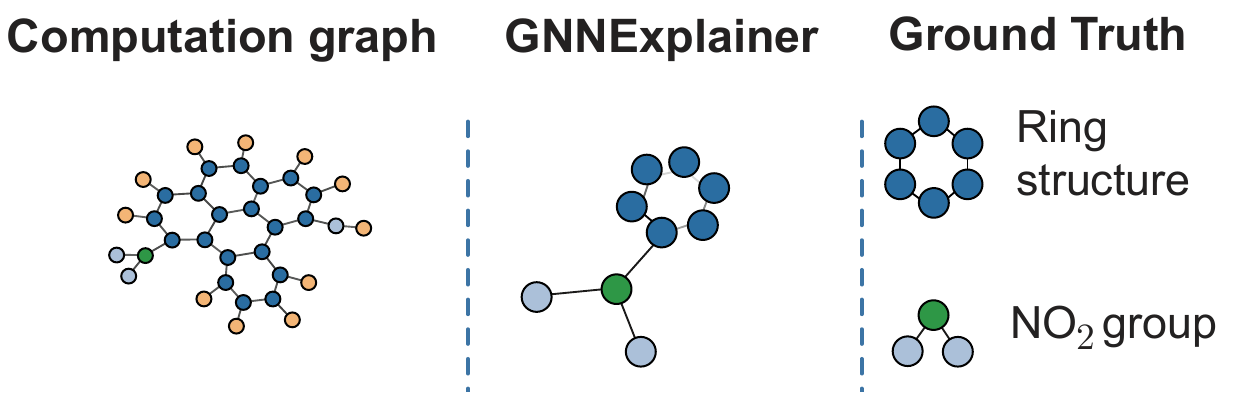}}}
\caption{ GNNExplainer generates an explanation by identifying a small graph of the input graph for  graph classification task  on  molecule graphs dataset (MUTAG). (Image Credit: ~\cite{ying2019gnnexplainer}) }
\label{fig:GNNExplainer}
\end{figure}

\begin{figure}[tb]
\centering
\centering
{\includegraphics[width=0.59\linewidth]{{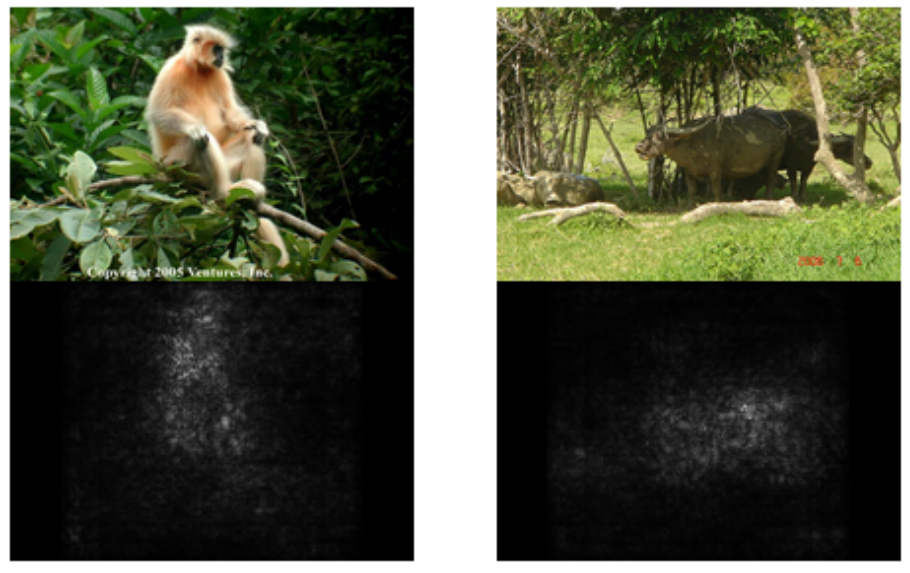}}}
\caption{  Image-specific class saliency maps were extracted using a single back-propagation pass through a DNN classification model. (Image Credit: ~\cite{simonyan2013deep}) }
\label{fig:SMV}
\end{figure}

\item \textbf{Model-agnostic Explanations}

The methods in this category are concerned with black-box well-trained AI models. More specifically, such methods do not try to create interpretable models, but to interpret already well-trained models. Such methods are widely used for explaining complicated models, such as deep neural networks. That is also why they sometimes are referred to as post hoc or black-box explainability methods in the related scientific literature. The great advantage of model-agnostic explanation methods over model-intrinsic ones is their flexibility.  Model-Agnostic methods are also widely applied in a variety of input modalities, such as images, text, graph-structured data, etc. Note that model-agnostic methods can also be applied to intrinsically interpretable models.

One of the most representative works in this category is Local Interpretable Model-Agnostic Explanations (\textbf{LIME})~\cite{ribeiro2016should}. For example, at the image domain, for any trained classifier, LIME is a proxy approach to randomly permute data by identifying the importance of local contiguous patches with similar pixels in a given instance and its corresponding label~\cite{ribeiro2016should}.  An illustrative example of LIME on a single instance for the top three predicted classes is shown in Figure~\ref{fig:lime}.

Additionally, to understand how any graph neural networks (GNNs) make a certain decision on graph-structured data, GNNExplainer learns soft masks for edges and node features to explain the predictions via maximizing the mutual information between the predictions of the original graph and those of the newly obtained graph~\cite{ying2019gnnexplainer,luo2020parameterized}. Figure~\ref{fig:GNNExplainer} illustrates explanation examples generated by GNNExplainer for graph-structured data.

\end{itemize}

\begin{figure}[tb]
\centering
\centering
{\includegraphics[width=0.693\linewidth]{{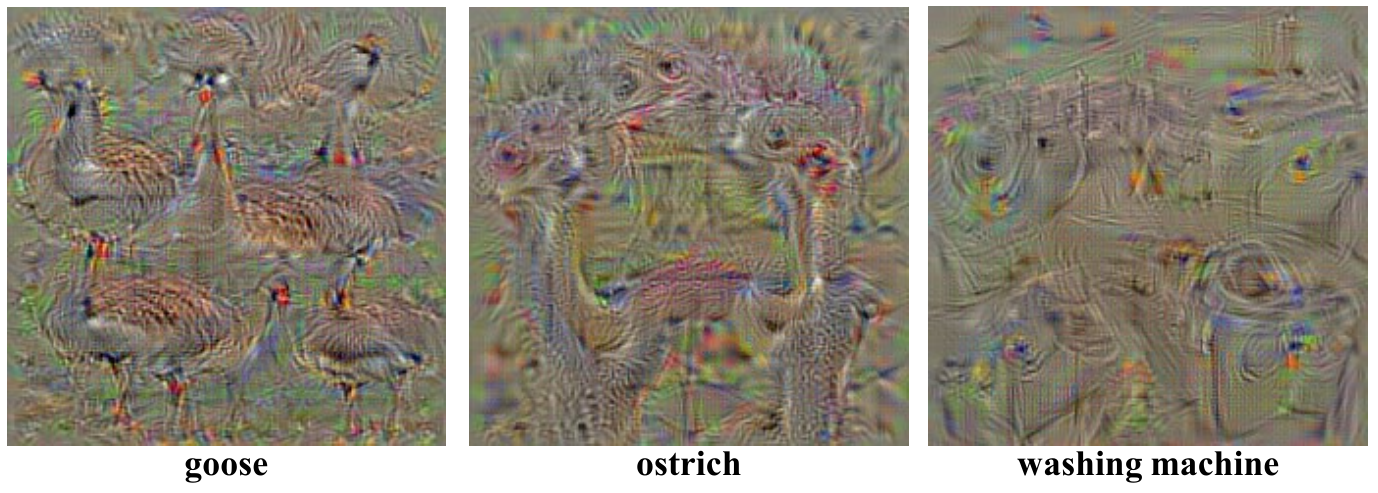}}}
\caption{  Numerically computed images, illustrating the class appearance models. (Image Credit: ~\cite{simonyan2013deep}) }
\label{fig:SMV2}
\end{figure}

\subsubsection{Scope of Explanation}

One important aspect of dividing the explainability techniques is based on the scope of explanation, i.e.,  local or global.

\begin{itemize}
\item \textbf{Local Explanations}

In general, the goal of locally explainable methods is to express the individual feature attributions of a single instance of input data $x$ from the data population $X$.  
For example, given a text document and a model to understand the sentiment of text, a locally explainable model might generate attribution scores for individual words in the text. 

In the Saliency Map Visualization method~\cite{simonyan2013deep}, the authors compute the gradient of the output class category with regard to an input image. By visualizing the gradients, a fair summary of pixel importance can be achieved by studying the positive gradients that have more influence on the output. An example of the class model is shown in Figure~\ref{fig:SMV}.

\item \textbf{Global Explanations}

The goal of global explanations is to provide insights into the decision of the model as a whole and to have an understanding about attributions for a batch of input data or a certain label, not just for individual inputs. In general, globally explainable methods work on an array of inputs to summarize the overall behavior of the black-box model. Most linear, rule-based and tree-based models are inherently globally explainable.  For example, conditional statements (intermediate nodes) in decision trees can give insight into how the model behaves in a global view, as shown in Figure~\ref{fig:dt}.

In terms of the DNNs models, Class Model Visualization~\cite{simonyan2013deep}  is trying to generate a particular image visualization by maximizing the score of class probability with respect to the input image.
An example of the class model is shown in Figure~\ref{fig:SMV2}.
\end{itemize}

\begin{figure}[tb]
\centering
\centering
{\includegraphics[width=0.65\linewidth]{{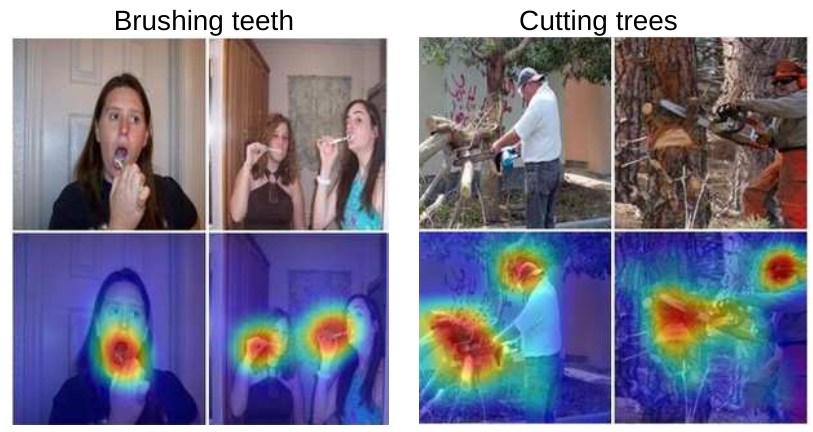}}}
\caption{Gradient-based Explanation: the CAM model produces class-specific regions of target images for visual explanation. (Image Credit: ~\cite{zhou2016learning}) }
\label{fig:CAM}
\end{figure}

\begin{figure}[tb]
\centering
\centering
{\includegraphics[width=0.693\linewidth]{{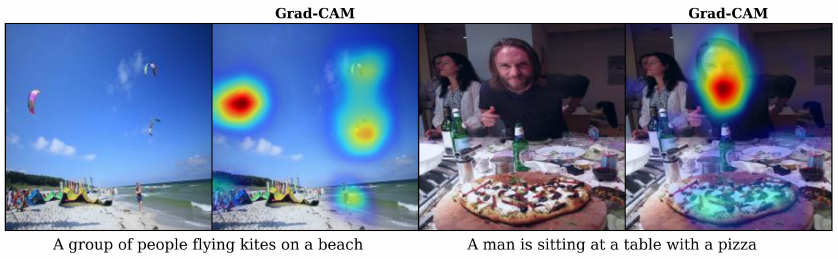}}}
\caption{ Gradient-based Explanation: the Grad-CAM model localizes image regions considered to be important for producing the captions. (Image Credit: ~\cite{selvaraju2017grad}) }
\label{fig:Grad-CAM}
\end{figure}

\subsubsection{Differences in the methodology}

This category is mainly defined by answering the question, "What is the algorithmic approach? Does it focus on the input data instance or the model parameters?" Based on the core algorithmic approach of the explanation method, we can categorize explanation methods as the ones that focus on the gradients of the target prediction with respect to input data, and those that focus on the changes or modifications of input data. 

\begin{itemize}
\item \textbf{Gradient-based Explanations}

In gradient-based methods, the explainable algorithm does one or more forward passes through the neural networks and generates attributions during the back-propagation stage utilizing partial derivatives of the activations.
This method is the most straightforward solution and has been widely used in computer vision to generate human-understandable visual explanations.

To understand how a CNN model makes decisions, Class activation mapping (CAM)~\cite{zhou2016learning} proposes modifying fully connected layers of the original CNN architecture using Global Average Pooling and generating important class-specific regions of the image for visual explanations via forward passes process. 
An illustrative example is provided in Figure~\ref{fig:CAM}.
Afterward, Gradient-weighted Class Activation Mapping (Grad-CAM)~\cite{selvaraju2017grad} generalizes the CAM model for any CNN model without requiring architectural changes or retraining and utilizes the gradient signal flowing into the final convolutional layer of a CNN for highlighting the important regions in the image. Figure~\ref{fig:Grad-CAM} shows a visual explanation via Grad-CAM for an image captioning model. 

\item  \textbf{Perturbation-based Explanations}

Perturbation-based explainable methods focus on variations in the input feature space to explain individual feature attributions toward the output class. More specifically,  explanations are generated by iteratively probing a trained AI model with different variations of the inputs. 
These perturbations can be on a feature level by replacing certain features with zero or random counterfactual instances, picking one or a group of pixels (super-pixels) for explanation, blurring, shifting, or masking operations, etc. In general, only forward pass is sufficient to generate the attribution representations without the need for back-propagating gradients. Shapley Additive explanations (SHAP)~\cite{lundberg2017unified} visualizes feature interactions and feature importance by probing feature correlations by removing features in a game-theoretic framework.

\end{itemize}

\subsubsection{Explanation via Other Approaches: Counterfactual Explanations}

In general, counterfactual explanations have been designed to answer hypothetical questions and describe how altering feature values of an instance would change the prediction to a predefined output~\cite{molnar2020interpretable,mittelstadt2019explaining}. 
Taking credit card application as an example, Peter gets rejected by AI banking systems and wonders why his application was rejected. To answer the question of "why," counterfactual explanations can be formulated as "What would have happened to this decision (from rejected to approved), if performing minimal changes in feature values (e.g., income, age, race, etc.)?" In fact, counterfactual explanations are usually human-friendly, since they are contrastive to the current instances and usually focus on a small number of features. To generate counterfactual explanations, Wachter et. al~\cite{wachter2017counterfactual} propose Lagrangian style constrained optimization as follows:
\begin{align}
\arg \min_{\hat{\mathbf{x}}} \max_{\lambda} \lambda \cdot (f(\hat{\mathbf{x}}) - \hat{y})^2 +d(\hat{\mathbf{x}}, \mathbf{x})
\end{align}
where $\mathbf{x}$ is the original instance feature, and $\hat{\mathbf{x}}$ is the corresponding counterfactual input. $f(\hat{\mathbf{x}})$ is the predicted result of a classifier.  The first term is the quadratic distance between the model prediction for the counterfactual $\hat{\mathbf{x}}$ and the targeted output $\hat{y}$. The second term indicates  the distance $d(\cdot, \cdot)$ between the instance $\mathbf{x}$ to be explained and the counterfactual $\hat{\mathbf{x}}$, and $\lambda$ is proposed to achieve the trade-off between the distance in prediction and the distance in feature values.

\begin{table*}[ht]
\centering
\caption{Summary of Published Research in Explainability of AI Systems. }
\label{tab:XAI}
\begin{tabular}{|c|c|l|c|l|l|c|l|}
\hline
\textbf{Representative Models} & \multicolumn{2}{c|}{\textbf{Model Usage}}                                                     & \multicolumn{3}{c|}{\textbf{Scope}}                                                                            & \multicolumn{2}{c|}{\textbf{Methodology}}                                                   \\ \hline
Linear model  & \multicolumn{2}{c|}{Intrinsic}                                                                & \multicolumn{3}{c|}{Global}                                                                                    & \multicolumn{2}{c|}{-}                                                                      \\ \hline
LIME~\cite{ribeiro2016should}                      & \multicolumn{2}{c|}{Agnostic}                                                                 & \multicolumn{3}{c|}{Both}                                                                                      & \multicolumn{2}{c|}{Perturbation}                                                           \\ \hline 
CAM~\cite{zhou2016learning}                        & \multicolumn{2}{c|}{Agnostic}                                                                 & \multicolumn{3}{c|}{Local}                                                                                     & \multicolumn{2}{c|}{Gradient}                                                               \\ \hline
Grad-CAM~\cite{selvaraju2017grad}                  & \multicolumn{2}{c|}{Agnostic}                                                                 & \multicolumn{3}{c|}{Local}                                                                                     & \multicolumn{2}{c|}{Gradient}                                                               \\ \hline
SHAP~\cite{lundberg2017unified}                    & \multicolumn{2}{c|}{Agnostic}                                                                 & \multicolumn{3}{c|}{Both}                                                                                      & \multicolumn{2}{c|}{Perturbation}                                                           \\ \hline
Saliency Map Visualization~\cite{simonyan2013deep} & \multicolumn{2}{c|}{Agnostic}                                                                 & \multicolumn{3}{c|}{Local}                                                                                     & \multicolumn{2}{c|}{Gradient}                                                               \\ \hline
GNNExplainer~\cite{ying2019gnnexplainer}           & \multicolumn{2}{c|}{Agnostic}                                                                 & \multicolumn{3}{c|}{Local}                                                                                     & \multicolumn{2}{c|}{Perturbation}                                                               \\ \hline
Class Model Visualization~\cite{simonyan2013deep}  & \multicolumn{2}{c|}{Agnostic}                                                                 & \multicolumn{3}{c|}{Global}                                                                                     & \multicolumn{2}{c|}{Gradient}                                                               \\ \hline
\textbf{Surveys} & \multicolumn{7}{c|}{~\cite{doshi2017towards,guidotti2018survey,zhang2018explainable,miller2019explanation,du2019techniques,molnar2020interpretable,belle2020principles,tjoa2020survey,arrieta2020explainable,yuan2020explainability,jimenez2020drug,danilevsky2020survey,linardatos2021explainable}} \\ \hline
\end{tabular}
\end{table*}

\subsection{Applications in Real Systems}
In this subsection, we discuss representative real-world applications where explainability is crucial. 

\subsubsection{Recommender Systems}
Recommender systems (RecSys) have become increasingly important in our daily lives since they play an important role in mitigating the information overload problem~\cite{fan2019graph,fan2020graph}. These systems provide personalized information to help human decisions and have been widely used in various user-oriented online services~\cite{fan2021attacking}, such as e-commerce item recommendations for everyday shopping (e.g., Amazon, Taobao), job recommendations for employment markets (e.g., LinkedIn), and friend recommendations to make people better connected (e.g., Facebook, Weibo)~\cite{fan2019deep,fan2019deep_daso}. Recent years have witnessed the great development of deep learning-based recommendation models, in terms of the improving accuracy and broader application scenarios~\cite{fan2019deep}. Thus, increasing attention has been paid to understanding why certain items have been recommended by deep learning-based recommender systems for end users, because providing good explanations of personalized recommender systems can sufficiently motivate users to interact with items, help users make better and/or faster decisions, and increase users' trust in the intelligent recommender systems~\cite{ma2019jointly,zhang2014explicit}. For example, to achieve explainability in recommender systems, RuleRec~\cite{ma2019jointly} proposes a joint learning framework for accurate and explainable recommendations by integrating induction of several explainable rules from item association, such as, \emph{Also view, Buy after view, Also buy, and Buy together}. 
The work~\cite{wang2018tem} proposes a tree-enhanced embedding method that seamlessly combines embedding-based methods with decision tree-based approaches, where a gradient boosting decision trees (GBDT) and an easy-to-interpret attention network are introduced to make the recommendation process fully transparent and explainable from  the side information of users and items.

\subsubsection{Drug discovery}
In the past few years, explainable AI has been proven to significantly accelerate the process of computer-assisted drug discovery~\cite{vamathevan2019applications,jimenez2020drug}, such as molecular design, chemical synthesis planning, protein structure prediction, and macromolecular target identification.
For example, explanations of graph neural networks have been conducted on a set of molecules graph-labeled 
for their mutagenic effect on the  Gram-negative bacterium Salmonella typhimurium, with the goal of identifying several known mutagenic functional groups $NH_2$ and $NO_2$~\cite{ying2019gnnexplainer,luo2020parameterized,yuan2020xgnn}.
A recent work~\cite{preuer2019interpretable} studies how the interpretation of filters within message-passing networks can lead to the identification of relevant toxicophore- and pharmacophore-like sub-structures for explainability, so as to help increase their reliability and foster their acceptance and usage in drug discovery and medicinal chemistry projects.

\subsubsection{Natural Language Processing (NLP)}
As one of the most broadly applied areas of AI, Natural Language Processing (NLP) investigates the use of computers to process or to understand human (i.e., natural) languages~\cite{deng2018deep}.
Applications of NLP are everywhere, including dialogue systems, text summarization, machine translation, question answering, sentiment analysis, information retrieval, etc. Recently, deep learning approaches have obtained very promising performance across many different NLP tasks, which comes at the expense of models becoming less explainable~\cite{danilevsky2020survey,mullenbach2018explainable}. To address the issue, CAML~\cite{mullenbach2018explainable}  employs an attention mechanism to  select the segments that are most relevant for medical codes (ICD) from clinical text.
LIME~\cite{ribeiro2016should} proposes to generate random input perturbations for a given document to explain the predicted categories for text classification in SVM models.

\subsection{Surveys and Tools}
In this subsection, we introduce existing surveys, tools and repositories on explainability in AI to facilitate the readers who wish to further explore this field.

\subsubsection{Surveys}

In the book~\cite{molnar2020interpretable}, the author focuses on interpretable machine learning by introducing from fundamental concepts to advanced interpretable models. For example, it first details related concepts of interpretability, followed by intrinsically interpretable models, such as linear regression, decision tree, rule-based methods, etc. Afterward, the book provides general model-agnostic tools for interpreting black-box models and explaining individual predictions. 
Doshi-Velez et al.~\cite{doshi2017towards} raises the importance of intractability in machine learning and introduces a comprehensive survey at this field. There are surveys ~\cite{gilpin2018explaining,belle2020principles,guidotti2018survey,linardatos2021explainable,du2019techniques,arrieta2020explainable} summarizing explanation approaches in machine learning. In addition, comprehensive surveys for specific applications also exist, such as recommender systems~\cite{zhang2018explainable}, medical information systems~\cite{tjoa2020survey}, natural language processing~\cite{danilevsky2020survey}, graph neural networks~\cite{yuan2020explainability}, Drug discovery~\cite{jimenez2020drug}, etc.

\subsubsection{Tools}
\label{sec:inter_tool}
In this subsection, we introduce several popular toolkits that are open-sourced in the GitHub platform for explainable AI. \textit{AIX360}\footnote{\url{https://aix360.mybluemix.net}} (AI Explainability 360)~\cite{arya2020ai} is an open-source Python toolkit featuring state-of-the-art explainability methods and some evaluation metrics. Meanwhile, AIX360 also provides educational materials for non-technical stakeholders to quickly become familiar with interpretation and explanation methods. \textit{InterpretML}\footnote{\url{https://github.com/interpretml/interpret}}~\cite{nori2019interpretml} is also an open-source python toolkit that exposes machine learning interpretability algorithms to practitioners and researchers. InterpretML exposes two types of interpretability – glassbox for machine learning models with model-intrinsic explanations, and black-box explainability techniques for explaining any existing AI systems. The package \textit{DeepExplain}~\cite{ancona2018towards} mainly supports various gradient-based techniques and perturbation-based methods\footnote{\url{https://github.com/marcoancona/DeepExplain}}. In addition, \textit{DIG}~\cite{liu2021dig} provides python toolkit for explaining graph deep learning.\footnote{\url{https://github.com/divelab/DIG}}

\subsection{Future Directions}

 In this subsection, we discuss potential directions for future research in explainable AI. Since the interpretability of AI is a relatively new and still a developing area, many open problems need to be considered.

\begin{itemize}
\item \textbf{Security of explainable AI.}  Recent studies have demonstrated that due to their data-driven nature, explanations of AI models are vulnerable to malicious manipulations. Attackers attempt to generate adversarial examples that can not only can mislead a target classifier but also can deceive its corresponding interpreter~\cite{zhang2020interpretable,ghorbani2019interpretation}, naturally raising potential security concerns on interpretations. Therefore, learning how to defend against adversarial attacks on explanations would be an important future direction for research.

\item \textbf{Evaluation Methodologies.}
Evaluation metrics are crucial for studying explanation methods; however, due to the lack of ground truths and human subjective understandings, evaluating whether the explanations are reasonable and correct in regard to certain predictions is becoming intractable. The widely used evaluation methodology is based on human evaluations based on visualizing explanations, which is time-consuming and biased toward subjective human understandings.  Although there are some initial studies on the evaluation of interpretability~\cite{doshi2017towards}, it is still unclear how to measure what constitutes a good Explanation?". It is crucial to investigate qualitative and quantitative evaluations of interpretability.

\item \textbf{Knowledge to Target model: from white-box to black-box.} Most existing explanation techniques require full knowledge of the explained AI system (denoted as white-box).  However, knowledge regarding target AI systems is often limited in many scenarios due to privacy and security concerns.   Therefore, an important direction is to understand is how an explanation can be generated for making decisions in black-box systems. 
\end{itemize}

\section{Privacy}
\label{sec:privacy}

The success of modern AI systems is built upon data, and data might contain private and sensitive information --  from credit card data to medical records and from social relations to family trees. To establish trustworthy AI systems, the safety of private and sensitive information carried by the data and models that could be potentially exposed throughout the AI system must be guaranteed. Therefore, increasing attention has been paid to the protection and regulation of data privacy.
From a legal perspective, laws from the state level to the global level have begun to provide mandatory regulations for data privacy. For instance, the California Consumer Privacy Act (CCPA) was signed into law in 2018 to enhance privacy rights and consumer protection in California by giving consumers more control over the personal information that businesses collect;
the Health Insurance Portability and Accountability Act (HIPAA) was created in 1996 to protect individual healthcare information by requiring authorization before disclosing personal healthcare information; 
the European Union announced General Data Protection Regulation (GDPR) to protect data privacy by giving the individual control over the personal data collection and usage. 

From the perspective of science and technology, although most AI technologies haven't considered privacy as the fundamental merit when they are first developed, to make modern AI systems trustworthy in privacy protection,
a subfield of AI, privacy-preserving machine learning (PPML), has set privacy protection as the priority and has begun to pioneer principled approaches for preserving privacy in machine learning. Specifically, researchers uncover the vulnerabilities of existing AI systems from comprehensive studies and then develop promising technologies to mitigate these vulnerabilities. In this section, we will provide a summary of this promising and important field. Specifically, the basic concepts and taxonomy will be first discussed, and the risk of privacy breaches will be explained through various privacy attacking methods. Mainstream privacy-preserving technologies, such as \textit{confidential computing}, \textit{federated learning}, and \textit{differential privacy} will be included, followed by discussions on applications in real systems, existing surveys and tools, and the future directions.

\subsection{Concepts and Taxonomy}
 
In the context of privacy protection, the adversarial goal of an attacker is to extract information about the data or machine learning models. According to the accessible information the adversary has, the attacker can be categorized into \textit{white-box or black-box}. In a white-box setting, we assume that the attacker has all information except the data that we try to protect and the attacker aims to attack. In a black-box setting, the attacker has very limited information, for example, the query results returned by the model. 
Based on when the attack occurs, the privacy breach could happen in the \textit{training phase or inference phase}. In the training phase, the adversary might be able to directly access or infer the information about the training data when she inspects or even tampers with the training process. In the inference phase, the adversary might infer the input data of the model by inspecting the output characteristics. 
According to the capability of the adversary, the attacker may be \textit{honest-but-curious or fully malicious}. An honest-but-curious attacker can inspect and monitor the training process while a fully malicious attacker can further tamper the training process. These taxonomies are not exclusive since they view the attacker from different perspectives.

\subsection{Methods}
We will highlight the risk of privacy leakage by introducing some representative privacy attack methods. Then, some mainstream techniques for privacy-preserving will be introduced.

\subsubsection{Privacy Attack}
Privacy attacks can target training data, input data, properties of data population, even the machine learning model itself. We introduce some representative privacy attacks to reveal the risk of privacy breaches.

\textbf{Membership Inference Attack.}
To investigate how machine learning models leak information about individual data within the training data, the membership inference attack aims to identify whether a data record is used in the training of model learning models. For instance, given the black-box access to the model, an inference model can be trained to recognize the differences of a target model's predictions on the inputs that are used in its training or not~\citep{shokri2017membership}. Empirically, it is shown that commonly used classification models can be vulnerable to membership inference attacks. Therefore, private information can be inferred if some user data (e.g., medical record and credit card data) is used in training the model. Please refer to the survey~\citep{hu2021membership} for a comprehensive summary on membership inference attack.

\textbf{Model Inversion Attack.}
Model inversion attack~\citep{fredrikson2014privacy, fredrikson2015model} aims to use the model's output to infer the information of the input data that often contains sensitive and private information. For instance, in pharmacogenetics, machine learning models are used to guide medical treatments, given the patient’s genotype and demographic information. However, it has been shown that severe privacy risk exists because a patient's genetic information can be disclosed given the model and the patient's demographic information~\citep{fredrikson2014privacy}. In facial recognition with neural networks, the images of people's faces can be recovered given their names, prediction confidence values, and access to the model~\citep{fredrikson2015model}. In~\citep{zhang2020secret}, generative adversarial networks (GANs) are used to guide the inversion process of neural networks and reconstruct high-quality facial images from face recognition classifiers. In a recent study, researchers found that the input data can be perfectly recovered through the gradient information of neural networks~\citep{zhu2020deep}, which highlights the privacy risk in distributed learning where gradient information needs to be transmitted when people used to believe that it can preserve data privacy.

\textbf{Property Inference Attack.}
Given the machine learning model, the property inference attack aims to extract global properties of the training dataset or training algorithm that the machine learning models do not intend to share.
One example is to infer the properties that only hold for a subset of the training data or a specific class of the training data. This type of attack might leak private statistical information about the population and the learned property can be used to exploit the vulnerability of an AI system. 

\textbf{Model Extraction.}
An adversary aims to extract the model information by querying the machine learning model in a black-box setting such that he can potentially fully reconstruct the model or create a substitute model that closely approximates the target model~\citep{tramer2016stealing}. Once the model has been extracted, the black-box setting translates to the white-box setting, where other types of privacy attacks become much easier. Moreover, the model information typically contains the intelligent property that should be kept confidential. For instance, ML-as-a-service (MLaaS) system, such as Amazon AWS Machine Learning, Microsoft Azure Machine Learning Studio, Google Cloud Machine Learning Engine, allow users to train the models on their data and provide publicly accessible query interfaces on a pay-per-query basis. The confidential model contains users' intelligent property but suffers from the risk of functionality stealing.

\subsubsection{Privacy Preservation}

\begin{figure}[h]
\includegraphics[width=0.45\textwidth]{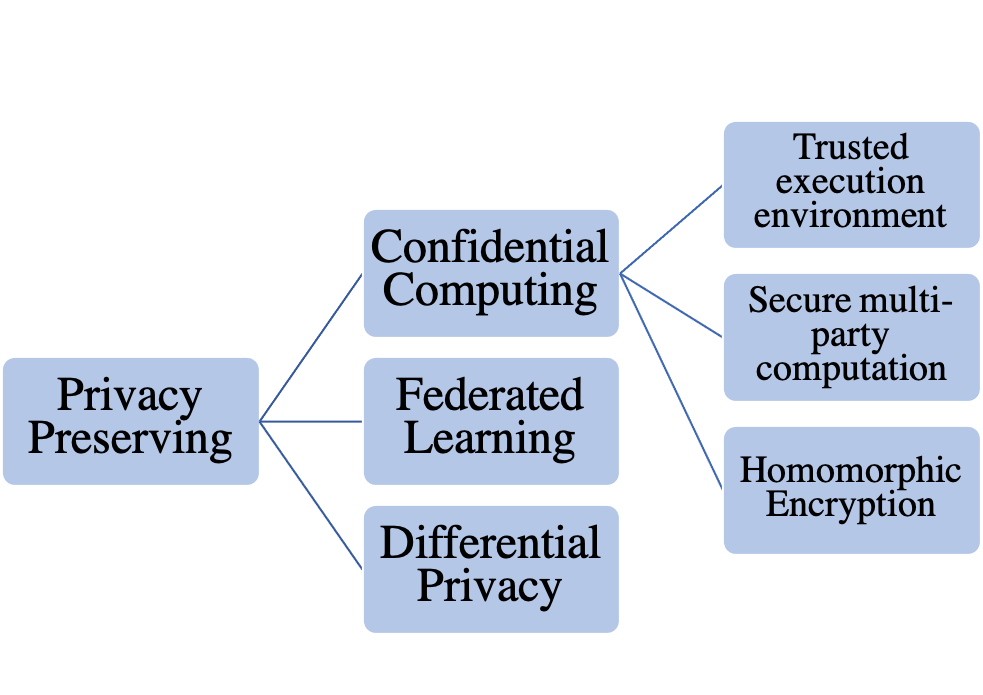}
\caption{An Overview of Privacy Preserving Techniques}
\label{fig:privacyoverview}
\end{figure}

The privacy-preserving countermeasures can be roughly categorized into three mainstream and promising directions, including confidential computing, federated learning, and differential privacy as shown in Figure~\ref{fig:privacyoverview}. Confidential computing attempts to ensure data safety during transmission and computing. Federated learning provides a new machine learning framework that allows data to be local and decentralized and avoid raw data transmission. Differential privacy aims to utilize the information about a whole dataset without exposing individual information in the dataset. Next, we review these techniques and discuss how they preserve privacy.

\textbf{Confidential Computing.}
There are mainly three types of techniques for achieving confidential computing, including Trusted Executive Environment (TEE)~\cite{sabt2015trusted}, Homomorphic Encryption (HE)~\cite{acar2018survey}, and Multi-party Secure Computation (MPC)~\cite{mpc-book}. 

\textit{Trusted Execution Environments.} 
Trusted Execution Environments focus on developing hardware and software techniques to provide an environment that isolates data and programs from the operator system, virtual machine manager, and other privileged processes. The data is stored in the trusted execution environment (TEE) such that it is impossible to disclose or operate on the data from outside. The TEE guarantees that only authorized codes can access the protected data, and the TEE will deny the operation if the code is altered. As defined by the Confidential Computing Consortium~\cite{ccc_whitepaper}, the TEE provides a level of assurance of data confidentiality, data integrity, and code integrity that essentially states that unauthorized entities cannot view, add, remove, or alter the data while it is in use within the TEE, and cannot add, remove or alter code executing in the TEE.

\textit{Secure Multi-party Computation.}
Secure multi-party computation (MPC) protocols aim to enable a group of data owners who might not trust one another to jointly perform a function computation that depends on all of their private input while without disclosing any participant's private data. Although the concept of secure computation was primarily a theoretical interest when it was first proposed~\cite{yao1982protocols}, it has now become a practical tool to enable privacy-preserving applications in which multiple distrusting data owners seek to compute a function cooperatively~\cite{mpc-book}. 

\textit{Homomorphic Encryption.}
Homomorphic Encryption (HE) enables computation functions on the data without accessing the plaintext by allowing mathematical operations to be performed on ciphertext without decryption. It returns the computation result in the encrypted form, which can be decrypted just as the computation is performed on the decrypted data. With partially homomorphic encryption schemes, only certain operations can be performed, which limits them to specialized problems that can be reduced as the supported operations. Fully-homomorphic encryption (FHE) schemes aim to provide support for a universal set of operations so that any finite function can be computed. The first FHE scheme was proposed by Gentry~\cite{gentry2009fully},and was based on lattice-based cryptography. There have been a lot of recent interests in implementing FHE schemes~\cite{gentry2011implementing, chillotti2016faster}, but to build a secure, deployable, scalable system using FHE is still challenging.

\begin{figure}[h]
\includegraphics[width=0.5\textwidth]{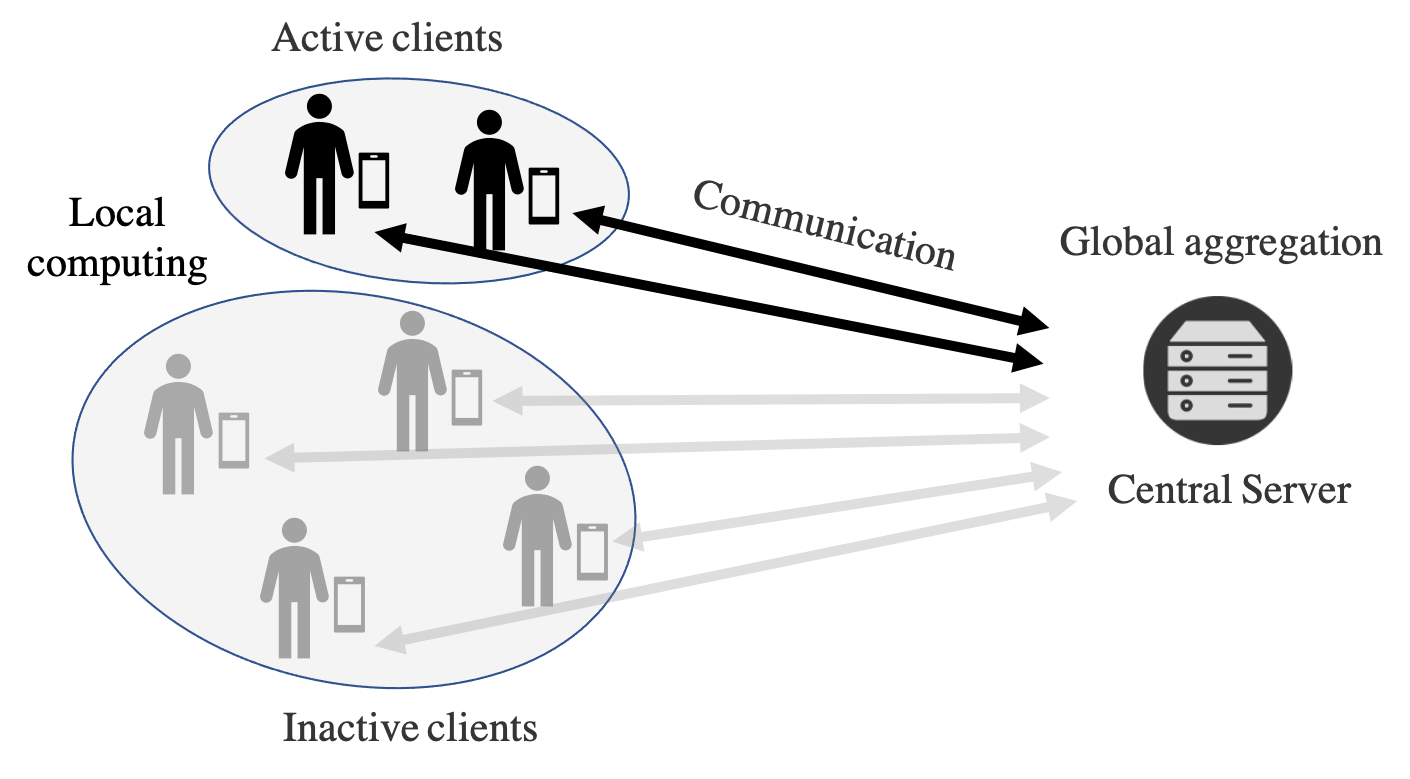}
\caption{Federated Learning}
\label{fig:fedml}
\end{figure}

\textbf{Federated Learning.}
Federated learning (FL), as shown in Figure~\ref{fig:fedml}, is a popular machine learning paradigm where many clients, such as mobile devices or sensors, collaboratively train machine learning models under the coordination of a central server, while keeping the training data from the clients decentralized~\citep{mcmahan2021advances}. This paradigm is in contrast with traditional machine learning settings where the data is first collected and transmitted to the central server for further processing. In federated learning, the machine learning models are moving between server and clients while keeping the private data locally within the clients. Therefore, it essentially avoids the transmission of private data and significantly reduces the risk of privacy breaches. 

Next, we briefly describe a typical workflow for a federated learning system~\cite{mcmahan2021advances}: 
\begin{itemize}
    \item \textit{Client selection}: The server samples a subset of clients from those active clients according to some eligibility requirements. 
    \item \textit{Broadcast}: The server broadcasts the current model and the training program to the selected clients. 
    \item \textit{Local computation}: The selected clients locally compute the update to the received model based on the local private data. For instance, the stochastic gradient descent (SGD) update can be run with the stochastic gradient computed based on local data and the model. 
    \item \textit{Aggregation}: The server collects the updated local models from the selected clients and aggregates them as an updated global model. 
\end{itemize}
This workflow represents one round of the federated learning algorithm and it will repeat until reaching specific requirements, such as the convergence accuracy or performance certificates.

In addition to protecting data privacy by keeping the data local, there are many other techniques to further secure data privacy. For instance, we can apply lossy compression before transferring the models between server and clients such that it is not easy for the adversary to infer accurate information from the model update~\cite{zhu2020deep}. We also can apply secure aggregation through secure multi-party computation such that no participant knows the local model information from the other participants, but the global model can still be computed~\cite{mohassel2017secureml, agrawal2019quotient}.
Additionally, we can also apply noisy perturbation to improve the differential privacy~\cite{wei2020federated}.

Federated learning is becoming an increasingly popular paradigm for privacy protection and has been studied, developed, and deployed in many applications. However, federated learning still faces many challenges, such as the efficiency and effectiveness of learning especially with non-IID data distributions~\cite{Li2020On, DBLP:conf/mlsys/LiSZSTS20, khaled2020tighter, karimireddy2020scaffold, liu2021linear}.

\textbf{Differential Privacy.} Differential Privacy (DP) is an area of research that aims to provide rigorous statistical guarantees for reducing the disclosure about individual information in a dataset~\cite{dwork2008differential, dwork2014algorithmic}. The major idea is to introduce some level of uncertainty through randomization or noise into the data such that the contribution of individual information is hidden while the algorithm can still leverage valuable information from the dataset as a whole. According to the definition~\cite{dwork2014algorithmic}, let's first define that the datasets $D$ and $D'$ are adjacent if $D'$ can be obtained from $D$ by altering the record of a single individual. A randomized algorithm $\mathcal{A}$ is $(\epsilon, \delta)$-differentially private if for all $\mathcal{S}\subset \text{Range}(\mathcal{A})$ and for all adjacent datasets $D$ and $D'$ such that
$$\text{Pr}[\mathcal{A}(D) \in \mathcal{S}] \leq e^{\epsilon} \text{Pr}(\mathcal{A}(D') \in \mathcal{S}) + \delta.$$
$(\epsilon, \delta)$ quantifies how much information can be inferred about an individual from the output of the algorithm $\mathcal{A}$ on the dataset. For instance, if $\epsilon$ and $\delta$ are sufficiently small, the output of the algorithm will be almost identical, i.e., $\text{Pr}[\mathcal{A}(D) \in \mathcal{S}] \approx \text{Pr}(\mathcal{A}(D') \in \mathcal{S})$, such that it is difficult for the adversary to infer the information of any individual since the individual's contribution on the output of the algorithm is nearly masked. The privacy loss incurred by the observation $\xi$ is defined as 
$$\mathcal{L}^\xi_{\mathcal{A}, D, D'} = \ln \Big ( \frac{\text{Pr}[\mathcal{A}[D]=\xi]}{\text{Pr}[\mathcal{A}[D']=\xi]} \Big ).$$ 
$(\epsilon, \delta)$-differential privacy ensures that for all adjacent datasets $D$ and $D'$, the absolute value of the privacy loss is bounded by $\epsilon$ with probability at least $1-\delta$. 
Some common methods to provide differential privacy include random response~\cite{warner1965randomized}, Gaussian mechanism~\cite{dwork2014algorithmic}, Laplace mechanism~\cite{dwork2006calibrating}, exponential mechanism~\cite{mcsherry2007mechanism}, etc.

\subsection{Applications in Real Systems}

Privacy-preserving techniques have been widely used to protect sensitive information in real systems. In this subsection, we discuss some representative examples.

\subsubsection{Healthcare} 

Healthcare data can be available from patients, clinical institutions, insurance companies, pharmacies, and so on. However, the privacy concern of personal healthcare information makes it difficult to fully exploit the large-scale and diverse healthcare data to develop effective predictive models for healthcare applications. Federated learning provides an effective privacy-preserving solution for such scenarios since data across the population can be utilized while not being shared~\citep{BRISIMI201859, kaissis2020secure, xu2021federated, rieke2020future, zerka2020systematic, sheller2020federated}. Differential privacy has also gained significant attention as a general way for protecting healthcare data~\citep{dankar2013practicing}.

\subsubsection{Biometric Data Analysis}

Biometric data is mostly non-revocable and can be used for identification and authentication. Therefore, it is critical to protect private biometric data. 
To this end, confidential computing, federated learning, and differential privacy techniques become widely applied to protect people's biometric data such as face image, medical image, and fingerprint pattern~\cite{bringer2013privacy, kaissis2020secure, sadeghi2009efficient, wang2020finprivacy}.

\subsubsection{Recommender Systems}
Recommender systems utilize users' interactions on products such as movies, music, and goods to provide relevant recommendations. The rating information has been shown to expose users to inference attacks, leaking private user attributes such as age, gender, etc~\cite{shyong2006you, aimeur2008lambic, mcsherry2009differentially}. To protect user privacy, recent works~\cite{mcsherry2009differentially, kapralov2013differentially, nikolaenko2013privacy, zhang2021graph} have developed privacy-preserving recommender systems via differential privacy.

\subsubsection{Distributed Learning}
In distributed learning, it is possible to recover a client's original data from their gradient information or model update~\cite{zhu2020deep, zhao2020idlg}. The Differentially private SGD~\cite{song2013stochastic, abadi2016deep, NEURIPS2020_fc4ddc15} provides an effective way to protect the input data by adding noise and has been popular in the training of deep learning models. Secure multiparty computing is applied to protect the information of locally trained models during aggregation~\cite{mohassel2017secureml, agrawal2019quotient, rouhani2018deepsecure}.

\subsection{Surveys and Tools}
\label{sec:pri_tool}
We collect some surveys and tools relevant to privacy in AI systems for further exploration.

\subsubsection{Surveys}
The general concepts, threats, attack and defense methods in privacy-preserving machine learning are summarized in several surveys~\citep{rigaki2020survey, al2019privacy, de2020overview}. 
Federated learning is comprehensively introduced in the papers~\citep{yang2019federated, mcmahan2021advances}. Differential privacy is reviewed in the surveys~\citep{dwork2008differential, dwork2014algorithmic, ji2014differential}.

\subsubsection{Tools}
Popular tools and repositories in federated learning include \textit{TensorFlow Federated (TFF)}~\cite{tensorflow_federated}, \textit{FATE}~\cite{fate}, \textit{FedML}~\cite{chaoyanghe2020fedml}, \textit{PaddleFL}~\cite{paddlefl} and \textit{LEAF}~\cite{leaf}. Popular tools in differential privacy include \textit{Facebook Opacus}~\cite{opacus}, \textit{TensorFlow-Privacy}~\cite{tf-privacy}, \textit{OpenDP}~\cite{opendp} and \textit{Diffpriv}~\cite{rubinsteindiffpriv}. 
\textit{Keystone Enclave}~\cite{lee2019keystone} is an open framework for designing Trusted Execution Environments. Popular tools in Secure Multiparty Computing and Homomorphic Encryption are summarized in the lists~\cite{mpc-list, he-list}.

\subsection{Future Directions}
Confidential computing, federated learning, and differential privacy are three effective ways to improve privacy protection. However, they are far away from being extensively used and require more development. For instance, the computation efficiency and flexibility of confidential computing are not mature enough to support the applications of AI systems in our society. There are also great challenges for improving the efficiency and effectiveness of federated learning when deploying in large-scale and heterogeneous environments. It would be desirable to achieve a better trade-off between utility and privacy loss in differential privacy. Most importantly, a versatile and reliable system design for achieving privacy protection and different techniques should be integrated to enhance the trustworthiness of AI systems.

\section{Accountability \& Auditability}
\label{sec:account}

In general, accountability for AI indicates how much we can trust these AI technologies and who or what we should blame if any parts of the AI technologies perform below expectation. It is about a declaration of responsibility. It is not trivial to explicitly determine the accountability for AI. On the one hand, most AI-based systems act as "black-box", due to the lack of explainability and transparency. On the other hand, real-world AI-based systems are very complex, and involve numerous key components, including input data, algorithm theory, implementation details, real-time human control, and so on. These factors further complicate the determination of accountability for AI. Although difficult and complex, it is necessary to guarantee accountability for AI. Auditability, which refers to a set of principled evaluations of the algorithm theories and implementation processes, is one of the most important methodologies in guaranteeing accountability,.

It is very important to achieve a balance between the accountability and innovation in AI. The overall aim is for humans to enjoy the benefits and conveniences of AI with a reliable and guarantee of safety. Additionally, however, we do not want to heavily burden the algorithm designer or put too many restrictions on end-users of AI-based systems. In this section, we discuss the accountability and auditability of AI. First, we introduce the basic concept of accountability and some key roles within it. We then describe the definition of auditability for AI and two kinds of audits. Finally, we summarize existing surveys and tools, and discuss some future directions to enhance the accountability and auditability in AI.

\subsection{Concepts and Taxonomy}
In this subsection, we will introduce the key concepts and taxonomies of accountability and auditability in AI. 

\subsubsection{Accountability}
Accountability in AI has a broad definition. On the one hand, accountability can be interpreted as a property of AI. From this perspective, accountability can be improved if breakthroughs can be made in the explainability of AI algorithms. On the other hand, accountability can be referred to as a clear responsibility distribution, which focuses on who should take the responsibility for each impact of AI-based systems. Here we mainly focus on discussing the second notion. As indicated above, it is not trivial to give a clear specification for responsibility, since the operation of an AI-based system involves many different parties, such as the system designer, the system deployer, and the end-user. Any improper operation from any parties may result in system failure or potential risk. Also, all kinds of possible cases should be taken into consideration to ensure a fair distribution of responsibility. For example, the cases when an AI system does harm when working correctly versus working incorrectly should be considered differently~\cite{martin2019ethical,yu2018building}. To better specify accountability, it is necessary to determine the roles and the corresponding responsibility of different parties in the function of an AI system. In~\cite{wieringa2020account}, three roles are proposed: decision-makers, developers, and users. By refining these three roles, we propose five roles, and introduce their responsibilities and obligations as follows: 
\textbf{System Designers}: system designers are the designers of the AI system. They are supposed to design an AI system that meets the user requirements and is transparent and explainable to the greatest extent. It is their responsibility to offer deployment instructions and user guidelines, and to release potential risks.
\textbf{Decision Makers}: decision-makers have the right to determine whether to build an AI system and what AI system should be adopted. Decision-makers should be fully aware of the benefits and risks of the candidate AI system, and take all the relevant requirements and regulations into overall consideration.
\textbf{System Deployers}: system deployers are in charge of deploying an AI system. They should follow the deployment instructions carefully and ensure that the system has been deployed appropriately. Also, they are expected to offer some hands-on tutorials to the end-users. 
\textbf{System Auditors}: system auditors are responsible for system auditing. They are expected to provide comprehensive and objective assessments of the AI system. 
\textbf{End Users}: end-users are the practical operators of an AI system. They are supposed to follow the user guidelines carefully and report emerging issues to system deployers and system designers in a timely fashion.

\subsubsection{Auditability} Auditability is one of the most important methodologies in ensuring accountability, which refers to a set of principled assessments from various aspects. In the IEEE standard for software development~\cite{ieee2018}, an audit is defined as ``an independent evaluation of conformance of
software products and processes to applicable regulations, standards, guidelines, plans, specifications, and procedures.” Typically, audits can be divided into two categories as follows:

\textbf{External audits}: external audits~\cite{green2019disparate,sandvig2014auditing}  refer to audits conducted by a third party that is independent of system designers and system deployers. External audits are expected to share no common interest with the internal workers and are likely to provide some novel perspectives for auditing the AI system. Therefore, it is expected that external audits can offer a comprehensive and objective audit report. However, there are obvious limitations to external audits. First, external audits typically cannot access all the important internal data in an AI system, such as the model training data and model implementation details~\cite{burrell2016machine}, which increases the auditing difficulty. Additionally, external audits are always conducted after an AI system is deployed, so that it may be costly to make adjustments over the system, and, sometimes, the system may have already done harm~\cite{moy2019police}.

\textbf{Internal audits}: internal audits refer to audits conducted by a group of people inside the system designer or system deployer organizations. SMACTR~\cite{raji2020closing} is a recent internal auditing framework proposed by researchers from Google and Partnership on AI and consists of five stages: scoping, mapping, artifact collection, testing, and reflection. Compared with external audits, internal audits can have access to a large amount of internal data, including the model training data and model implementation details, which makes internal audits much more convenient. Furthermore, internal audits can be conducted before an AI system is deployed, thus avoiding some potential harm after the system's deployment. The internal audit report can also serve as an important reference for the decision-maker to make a decision. However, an unavoidable shortcoming for internal audits is that they share the same interest as the audited party, which makes it challenging to give an objective audit report. 

\subsection{Surveys and Tools}
In this subsection, we summarize existing surveys and tools about accountability and auditability of AI, to facilitate readers who want to explore this field further.

\subsubsection{Surveys}
A recent work on algorithmic accountability is presented in~\cite{wieringa2020account}. It takes Boven's definition of accountability~\cite{bovens2007analysing} as the basic concept and combines it with numerous literature in algorithmic accountability to build the concept's definition.

\subsubsection{Tools}
The other five dimensions (safety \& robustness, non-discrimination \& fairness, explainability, privacy, environmental well-being) discussed in this survey are also important aspects to be evaluated during algorithm auditing. Therefore, most tools introduced in section ~\ref{sec:rob_repo}, \ref{sec:fair_tool}, \ref{sec:inter_tool}, \ref{sec:pri_tool}, and \ref{sec:env_tool} can also be used for the purpose of auditing.

\subsection{Future Directions}
For accountability, it is important to further enhance the explainability of the AI system. Only when we have a deep and thorough understanding of its theory and mechanism can we fully rely on it or make a well-recognized responsibility distribution scheme. For auditability, it is always a good option to conduct both external audits and internal audits, so that we can have a comprehensive and objective overview of an AI system. Furthermore, we need to be aware that an AI system is constantly dynamic. It can change with input data and environment. Thus, to make an effective and timely audit, it is necessary to audit the system periodically and to update auditing principles with the system changes~\cite{lins2019designing}.

\section{Environmental Well-being}
\label{sec:environment}

A trustworthy AI system should be sustainable and environmentally friendly \cite{smuha2019eu}. In fact, the large-scale development and deployment of AI systems brings a huge burden of energy consumption, which inevitably affects the environment. For example, Table \ref{table:carbon} shows the carbon emission (as an indicator of energy consumption) of training NLP models and that of daily consumption \cite{strubell2019energy}. We find that training a common NLP pipeline has the same carbon emissions as a human produces in seven years. Training and fine-tuning a large Transformer model costs five times more energy consumption than a car over its lifetime. Besides model development, in other areas, such as data center cooling,\footnote{\url{https://www.forbes.com/sites/forbestechcouncil/2020/08/17/why-we-should-care-about-the-environmental-impact-of-ai/?sh=b90512e56ee2}} there is also a huge energy cost. The rapid development of AI technology further challenges the tense global situation of energy shortage and environmental deterioration. Hence, environmental friendliness becomes an important issue to consider in building trustworthy AI systems. In this section, we review the existing works regarding the environmental impacts of AI technologies. Existing works mainly focus on the impact of AI systems' energy consumption on the environment. We first present an overview of the strategies for reducing energy consumption, e.g., model compression, and then introduce the works on estimating the energy consumption and evaluating the environmental impacts of real-world AI systems in different domains. Finally, we summarize the existing surveys and tools on this dimension.

\begin{table}[ht]
    \centering
    \begin{tabular}{lrr}
        \bf Consumption  & \bf CO$_\mathbf{2}$e (lbs)\\ \hline
        Air travel, 1 passenger, NY$\leftrightarrow$SF & 1984\\
        Human life, avg, 1 year & 11,023\\
        American life, avg, 1 year & 36,156 \\
        Car, avg incl. fuel, 1 lifetime & 126,000\\
        & & \\
        \bf Training one model (GPU)  & \\ \hline
        NLP pipeline (parsing, SRL) & 39 \\
        \ \ \ \ w/ tuning \& experimentation  & 78,468 \\
        Transformer (big)  & 192\\ 
        \ \ \ \ w/ neural architecture search & 626,155\\
    \end{tabular}
    \caption{Comparsions between estimated CO$_2$ emissions produced by daily lives and training NLP models. (Table Credit: \cite{strubell2019energy})}
    \label{table:carbon}
\end{table}

\subsection{Methods}
In this subsection, we summarize the techniques developed for reducing the energy use of AI algorithms. Improving the energy efficiency of AI systems involves algorithm-level and hardware-level solutions. We will introduce two common classes of algorithm-level approaches: model compression and adaptive design, as well as hardware-level energy-saving methods.

\subsubsection{Model Compression}
Model compression is a hot topic in deep learning that receives continuous attention from both academia and industry. It studies how to reduce the size of a deep model to save storage space and the energy consumption for training and deploying models, with an acceptable sacrifice on model performance. For CNN models in the image domain, parameter pruning and quantization \cite{han2015deep,choi2016towards}, low-rank factorization \cite{rigamonti2013learning,jaderberg2014speeding}, transferred/compact convolutional filters \cite{cohen2016group,shang2016understanding}, and knowledge distillation have been proposed \cite{hinton2015distilling,romero2014fitnets}. Similarly, in the text domain, researchers borrow and extend these methods: pruning \cite{cao2019efficient,michel2019sixteen}, quantization \cite{cheong2019transformers,hou2018loss}, knowledge distillation \cite{kim2016sequence,sun2019patient,tang2019distilling}, and parameter sharing \cite{dehghani2018universal,lan2019albert}, to compress popular NLP models, such as Transformer and BERT.

\subsubsection{Adaptive Design}
Another line of research focuses on adaptively designing a model architecture to optimize the energy efficiency of a model. 
\citet{yang2017designing} propose a pruning approach to design CNN architectures to achieve an energy-saving goal. In their method, the model is pruned in a layer-by-layer manner, where the layer that consumes the most energy is pruned first. \citet{stamoulis2018designing} propose a framework to adaptively design CNN models for image classification under energy consumption restrictions. They formulate the design of a CNN architecture as a hyperparameter optimization problem and solve it by Bayesian optimization. 

\subsubsection{Hardware}
In addition to the algorithm level, endeavors are also conducted to improve the energy efficiency of AI from the design of the hardware. Computing devices or platforms specially designed for AI applications are proposed to maximize the training and inference efficiency of AI algorithms. Specifically, hardware designed for DNN models are called DNN accelerators \cite{chen2020survey}. \citet{esmaeilzadeh2012neural} design a neural processing unit (NPU) to execute the fixed computations of a neuron such as multiplication, accumulation, and sigmoid, on chips. Later, \citet{liu2015reno} proposed RENO, which is a more advanced on-chip architecture for neural network acceleration. There is also hardware designed for specific NN structures. \citet{han2016eie} investigate how to design an efficient computation device for a sparse neural network, where weight matrices and feature maps are sparse. They \cite{han2017ese} also devise an efficient speech recognition engine that is dedicated to RNN models. Furthermore, ReGAN \cite{chen2018regan} is developed to accelerate generative adversarial networks (GANs).

\subsection{Applications in Real Systems}
As described before, the environmental impacts of AI systems mainly come from energy consumption. In this subsection, we introduce the research on evaluating and estimating the energy consumption of real-world AI systems in different domains.

In the field of computer vision, \citet{li2016evaluating} first investigate the energy use of CNNs on image classification tasks. They provide a detailed comparison among different types of CNN layers, and also analyze the impact of hardware on energy consumption. \citet{cai2017neuralpower} introduce the framework NeuralPower which can estimate the power and runtime across different layers in a CNN, to help developers to understand the energy efficiency of their models before deployment. They also propose evaluating CNN models with a novel metric ``energy-precision ratio''. Based on it, developers can trade off energy consumption and model performance according to their own needs, and choose the appropriate CNN architecture. In the field of NLP, \citet{strubell2019energy} examine the carbon emissions of training popular NLP models, namely, Transformer, ELMo, and GPT-2, on different types of hardware, and shed light on the potential environmental impacts of NLP research and applications.

\subsection{Surveys and Tools}
\label{sec:env_tool}
In this subsection, we collect related surveys and tools on the dimension of environmental well-being.

\subsubsection{Surveys}

From the algorithm-level perspective, \citet{garcia2019estimation} present a comprehensive survey on energy consumption estimation methods from both the computer architecture and machine learning communities. Mainly, they provide a taxonomy for the works in computer architecture and analyze the strengths and weaknesses of the methods in various categories. \citet{cheng2017survey} summarize the common model compression techniques and organize them into four categories, and then present detailed analysis on the performance, application scenarios, advantages, and disadvantages of each category. In the hardware-level perspective, \citet{wang2020benchmarking} compare the performance and energy consumption of the processors from different vendors for AI training. \citet{mittal2014survey} review the approaches for analyzing and improving GPU energy efficiency. The survey \cite{chen2020survey} summarizes the latest progress on DNN accelerator design.

\subsubsection{Tools}
\textit{SyNERGY} \cite{rodrigues2018synergy} is a framework integrated with Caffe for measuring and predicting the energy consumption of CNNs. \citet{lacoste2019quantifying} develop a \textit{Machine Learning Emissions Calculator} as a tool to quantitatively estimate the carbon emissions of training an ML model, which can enable researchers and practitioners to better understand the environmental impact caused by their models. \textit{Accelergy} \cite{wu2019accelergy} and \textit{Timeloop} \cite{parashar2019timeloop} are two representative energy estimation tools for DNN accelerators.

\subsection{Future Directions}
Research on reducing the energy consumption of AI systems for environmental well-being is on the rise. At the algorithmic level, automated machine learning (AutoML), which aims to automatically design effective and efficient model architectures for certain tasks, emerges as a novel direction in the AI community. Existing works in AutoML focus more on designing an algorithm to improve its performance, but don't usually treat energy consumption savings as the highest priority. Using AutoML technologies to design energy-saving models needs further exploration in the future. At the hardware level, current research on DNN accelerators pays more attention to devising efficient deployment devices to facilitate model inference, but the procedure of model training is overlooked. The design of efficient customized training devices for various DNN models is a practical and promising direction to investigate in the future.
\section{Interactions among Different Dimensions}
\label{sec:relation}
An ideal trustworthy AI system should simultaneously satisfy the six dimensions discussed above. In reality, the six dimensions are not independent of one another. The satisfaction of one dimension can promote the pursuit of another dimension. Meanwhile, conflicts exist among different dimensions. The realization of one dimension could violate another dimension, which makes it impossible for two or more dimensions to be met simultaneously in some scenarios. Researchers and practitioners should be aware of the complicated interactions among different dimensions. Knowing the accordance between two dimensions brings us an alternative idea to achieve one dimension: we can try to satisfy one dimension by realizing the other. Moreover, when two dimensions are contradictory, we can make a trade-off between them according to our needs. In this section, we discuss some known accordance and conflict interactions among different dimensions.

\subsection{Accordance} Two dimensions are accordant when the satisfaction of one dimension can facilitate the achievement of the other, or the two dimensions promote each other. Next, we show two examples of accordance interactions among dimensions. 

\textbf{Robustness \& Explainability.} Studies show that deep learning models' robustness against adversarial attacks positively correlates with their explainability. \citet{etmann2019connection} find that models trained with robustness objectives show more interpretable saliency maps. Specifically, they prove rigorously in mathematics that Lipschitz regularization, which is commonly used for robust training, forces the gradients to align with the inputs. \citet{noack2021empirical} further investigate the opposite problem: will an interpretable model be more robust? They propose Interpretation Regularization (IR) to train models with explainable gradients and empirically show that a model can be more robust to adversarial attacks if it is trained to produce explainable gradients.

\textbf{Fairness \& Environmental Well-being.}
Fairness in the field of AI is a broad topic, which involves not only the fairness of AI service providers and users, but also the equality of AI researchers. As mentioned in section~\ref{sec:environment}, the development trend of deep learning models toward larger models and more computing resource consumption not only causes adverse environmental impact but also aggravates the inequality of research~\cite{strubell2019energy}, since most researchers cannot afford high-performance computing devices. Hence, the efforts for ensuring the environmental well-being of AI techniques, such as reducing the cost of training large AI models, are in accordance with the fairness principle of trustworthy AI.

\subsection{Conflict}
Two dimensions are conflicting when the satisfaction of one dimension hinders the realization of the other. Next, we show three examples of the conflicting interactions among dimensions.

\textbf{Robustness \& Privacy.} Recent studies find tensions between the robustness and the privacy requirements of trustworthy AI. Song et al.~\cite{song2019privacy} checks how the robust training against adversarial attacks influences the risk of a model against membership inference attack. They find that models trained with adversarial defense approaches are more likely to expose sensitive information in training data via membership inference attacks. The reason behind this is that models trained to be robust to adversarial examples typically overfit to training data, which makes training data easier to be detected from models' outputs.

\textbf{Robustness \& Fairness.} Robustness and fairness can also conflict with each other in particular scenarios. As discussed in section~\ref{sec:safety}, adversarial training is one of the mainstream approaches for improving the robustness of a deep learning model. Recent research~\cite{xu2020robust} indicates that adversarial training can introduce a significant disparity between performance and robustness among different groups, even if the datasets are balanced. Thus, the adversarial training algorithm improves the robustness of a model at the expense of its fairness. Accordingly, the work~\cite{xu2020robust} proposes a framework called Fair-Robust-Learning (FRL) to ensure fairness while improving a model's robustness. 

\textbf{Fairness \& Privacy.} Cummings et al.~\cite{cummings2019compatibility} investigate the compatibility of fairness and privacy of classification models, and theoretically prove that differential privacy and exact fairness in terms of equal opportunity are unlikely to be achieved simultaneously. By relaxing the condition, this work further shows that it is possible to find a classifier that satisfies both differential privacy and approximate fairness.
\section{Future Directions}
\label{sec:future}

In this survey, we elaborated on six of the most concerning and crucial dimensions an AI system needs to meet to be trustworthy. Beyond that, some dimensions have not received extensive attention, but are worth exploring in the future. In this section, we will discuss several other potential dimensions of trustworthy AI.

\subsection{Human agency and oversight}
The ethical guidelines for trustworthy AI proposed by different countries and regions all emphasize the human autonomy principle of AI technology~\cite{smuha2019eu}. Human autonomy prohibits AI agents from subordinating, coercing, or manipulating humans, and requires humans to maintain self-determination over themselves. To achieve the principle of human autonomy, the design of AI systems should be human-centered. Specifically, human agency and oversight should be guaranteed in the development and deployment of AI systems. Human agency enables humans to make decisions independently based on the outputs of an AI system, instead of being totally subject to AI's decisions. A desirable human agency technology encourages users to understand the mechanism of an AI system and enables users to evaluate and challenge the decisions of an AI system, and make better choices by themselves. Human oversight enables humans to oversee AI systems throughout their life cycle, from design to usage. It can be achieved through human-in-the-loop, human-on-the-loop, and human-in-command governance strategies.

\subsection{Creditability}
With the wide deployment of AI systems, people increasingly rely on content produced or screened by AI, such as an answer to a question given by a question-answering (QA) agent or a piece of news delivered by a recommender system. However, the integrity of such content is not always guaranteed. For example, an AI system that exposes users to misinformation should not be considered trustworthy. Hence, additional mechanisms and approaches should be incorporated in AI systems to ensure their creditability.

\subsection{Interactions among Different Dimensions}
As discussed in section~\ref{sec:relation}, different dimensions of trustworthy AI can interact with one another in an accordant or conflicting manner. However, the research on the interactions among different dimensions is still in an early stage. Besides the several instances shown in this paper, there are potential interactions between other dimension pairs remaining to be investigated. For example, people may be interested in the relationship between fairness and interpretability. In addition, the interaction formed between two dimensions can be different in different scenarios, which needs more exploration. For example, an interpretable model may promote its fairness by making its decision process transparent. On the contrary, techniques to improve the interpretability of a model may introduce a disparity of interpretability among different groups, which leads to a fairness problem. Although there are numerous problems to study, understanding the interactions among different dimensions is very important in building a trustworthy AI system.
\section{Conclusion}
\label{sec:con}
In this survey, we present a comprehensive overview of trustworthy AI from a computational perspective and clarify the definition of trustworthy AI from multiple perspectives, distinguishing it from similar concepts. We introduce six of the most crucial dimensions that make an AI system trustworthy; namely, Safety \& Robustness, Nondiscrimination \& Fairness, Explainability, Accountability \& Auditability, Privacy, and Environmental Well-being. For each dimension, we present an overview of related concepts and a taxonomy to help readers understand, how each dimension is studied, and summarize the representative technologies, to enable readers to follow the latest research progress in each dimension. To further deepen the understanding of each dimension, we provide numerous examples of applications in real-world systems and summarize existing related surveys and tools. We also discuss potential future research directions within each dimension. We then analyze the accordance and conflicting interactions among different dimensions. Finally, it is important to mention that outside of the six dimensions elaborated in this survey, there still exist some other potential issues that may undermine our trust in AI systems. Hence, we discuss several possible dimensions of trustworthy AI as future research directions.

\bibliographystyle{ACM-Reference-Format}
\bibliography{main}

\end{document}